\documentclass[journal]{IEEEtran}

\usepackage{xcolor,soul,framed} 

\colorlet{shadecolor}{yellow}
\usepackage[pdftex]{graphicx}
\graphicspath{{../pdf/}{../jpeg/}}
\DeclareGraphicsExtensions{.pdf,.jpeg,.png}

\usepackage[cmex10]{amsmath}
\usepackage{array}
\usepackage{mdwmath}
\usepackage{mdwtab}
\usepackage{eqparbox}
\usepackage{url}
\usepackage{multirow}
\usepackage{amssymb}             
\usepackage{amsfonts}            
\usepackage{mathrsfs}            
\usepackage{booktabs}
\usepackage{tabularx}
\usepackage{adjustbox}

\usepackage{makecell}
\usepackage{hyperref}
\hypersetup{
colorlinks=true,
linkcolor=blue,
citecolor=blue
}

\usepackage[numbers,sort&compress]{natbib}
\begin{document}
\bstctlcite{IEEEexample:BSTcontrol}
    \title{IFViT:  Interpretable Fixed-Length Representation for Fingerprint Matching via Vision Transformer}
  \author{Yuhang Qiu, Honghui Chen, Xingbo Dong, Zheng Lin, Iman Yi Liao,~\IEEEmembership{Member,~IEEE}, \\ Massimo Tistarelli,~\IEEEmembership{Senior Member,~IEEE}, Zhe Jin,~\IEEEmembership{Member,~IEEE} \\
  {\thanks{This work was supported by the National Natural Science Foundation of China  (62376003) and Anhui Provincial Natural Science Foundation (2308085MF200) and High-end Foreign Experts Recruitment Plan (G2023019010L) by the Ministry of Science and Technology China. Yuhang Qiu, Xingbo Dong, and Zhe Jin are with the Anhui Provincial Key Laboratory of Artificial Intelligence, School of Artificial Intelligence, Anhui University, Hefei 230093, China. Yuhang Qiu is also with the Faculty of Engineering, Monash University, Wellington Road, Clayton, Victoria 3800, Australia (e-mail: yuhang.qiu@monash.edu; xingbo.dong@ahu.edu.cn; jinzhe@ahu.edu.cn) (Corresponding author: Zhe Jin).}
  \thanks{Honghui Chen is with the Department of Physics and Information Engineering, Fuzhou University, Fuzhou 350108, China (e-mail: 221110026@fzu.edu.cn).}
  \thanks{Zheng Lin is with the Department of Electrical and Electronic Engineering, University of Hong Kong, Pok Fu Lam, Hong Kong SAR, China (e-mail: linzheng@eee.hku.hk).} 
  \thanks{Iman Yi Liao is with the School of Computer Science, University of Nottingham Malaysia Campus, Semenyih 43500, Malaysia (e-mail: Iman.Liao@nottingham.edu.my).}
  \thanks{Massimo Tistarelli is with the Computer Vision Laboratory, University of Sassari, 07100 Sassari, Italy (e-mail: tista@uniss.it).}}
}



\maketitle

\begin{abstract}
Determining dense feature points on fingerprints used in constructing deep fixed-length representations for accurate matching, particularly at the pixel level, is of significant interest. To explore the interpretability of  fingerprint matching, we propose a multi-stage interpretable fingerprint matching network, namely Interpretable Fixed-length Representation for Fingerprint Matching via Vision Transformer (IFViT), which consists of two primary modules. The first module, an interpretable dense registration module, establishes a Vision Transformer (ViT)-based Siamese Network to capture long-range dependencies and the global context in fingerprint pairs. It provides interpretable dense pixel-wise correspondences of feature points for fingerprint alignment and enhances the interpretability in the subsequent matching stage. The second module takes into account both local and global representations of the aligned fingerprint pair to achieve an interpretable fixed-length representation extraction and matching. It employs the ViTs trained in the first module with the additional fully connected layer and retrains them to simultaneously produce the discriminative fixed-length representation and interpretable dense pixel-wise correspondences of feature points. Extensive experimental results on diverse publicly available fingerprint databases demonstrate that the proposed framework not only exhibits superior performance on dense registration and matching but also significantly promotes the interpretability in deep fixed-length representations-based fingerprint matching.


\end{abstract}

\begin{IEEEkeywords}
Interpretable Fingerprint Recognition, Vision Transformers, Fingerprint Registration and Matching, Fixed-Length Fingerprint Representation
\end{IEEEkeywords}

%
\IEEEpeerreviewmaketitle


\section{Introduction}

\begin{figure}
  \begin{center}
  \vspace{1 pt}
  \includegraphics[width=3.1in]{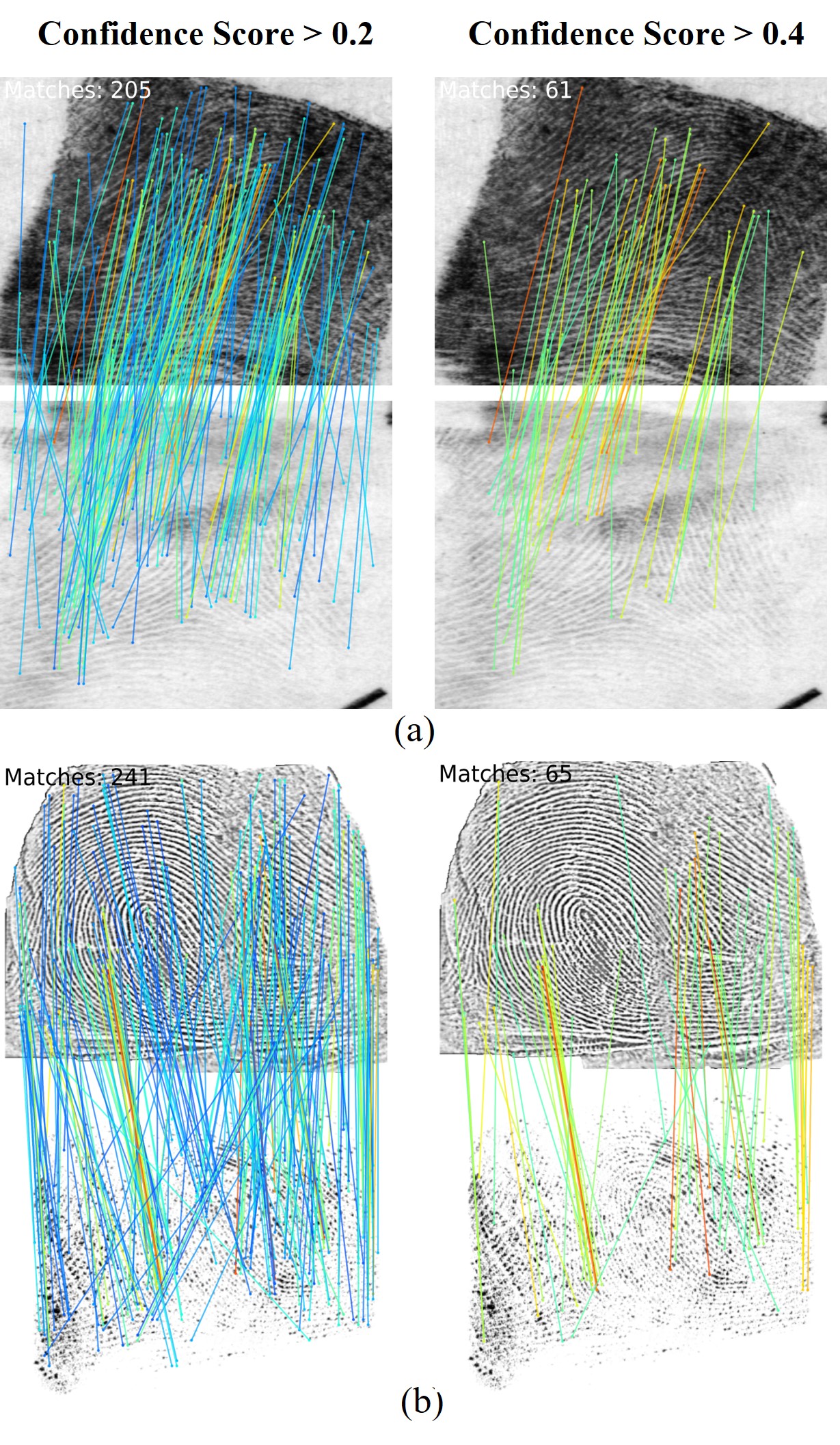}
  \caption{{ Produced dense pixel-wise correspondences of feature points from IFViT in the case of (a) low-quality and (b) cross-sensor fingerprint pairs selected with different thresholds of confidence scores.}}
  \vspace{-0.9cm} 
  \label{fig:Matching Imgs}
  \end{center}
\end{figure}

\IEEEPARstart{F}{ingerPrint} is an immutable and unique biological trait widely used for human authentication in various scenarios including forensics, bank identification and physical access control \cite{chhabra2020state,arora2020fingerprint}. As a crucial part of authentication, fingerprint matching aims to compare the input fingerprint patterns with those stored in a database to determine if they belong to the same finger. Minutiae, e.g. ridge endings and bifurcations, are commonly considered reliable features for accomplishing the matching process. However, extracting minutiae may be challenging when fingerprint quality is low due to conditions such as dry or wet \cite{engelsma2019learning}. Conversely, deep learning-based approaches are capable of extracting discriminative fixed-length fingerprint representation and have been considered a promising  alternative to address the limitations of traditional minutiae-based matching methods \cite{engelsma2019learning,grosz2021c2cl,cao2019endtoend,takahashi2020fingerprint}. Despite significant progress, the improvement of interpretability in deep learning-based fingerprint matching is still in its infancy.


Machine Learning (ML) methods have gained tremendous success in major fields due to their powerful inferential capabilities \cite{lin2023efficient, Qiu2024,yuan2023graph,lin2024adaptsfl,qiu2022pose}, among which explainable Artificial Intelligence (XAI) is currently one of the key focuses \cite{confalonieri2021historical, dobilovic2018explainable}. XAI aims to enhance the comprehensibility and transparency on the outcomes of artificial intelligence systems to facilitate reliable real-world data-driven applications. Understanding the underlying reasons behind ML decision-making is crucial, particularly for black-box deep learning models. For fingerprint matching, accurately determining feature points at the pixel level used for matching has the potential to significantly improve the interpretability. Although conventional minutiae-based matching approaches can establish pixel-wise correspondences of minutiae between an input fingerprint pair \cite{tico2003fingerprint}, performing pairwise comparisons of fingerprints is computation-intensive and varies with the number of detected minutiae, especially in the encrypted domain \cite{engelsma2019learning}. To overcome this challenge, learning fixed-length fingerprint representation has emerged as a promising solution  \cite{engelsma2019learning}. However, directly revealing pixel-wise correspondences of feature points within the provided fingerprint pair remains challenging.

Deep learning models have recently been explored to build correspondences for a given fingerprint pair. Cao et al. \cite{cao2019endtoend} introduced the Autoencoder-based model for minutiae detection and established minutiae correspondences for computing the similarity between latent and reference fingerprints. To mitigate the negative impact of nonlinear skin distortion,  fingerprint dense registration techniques \cite{cui2020dense, gu2020latent} are proposed to measure pixel-wise displacement between two fingerprints. However, such models primarily serve as the basis for subsequent tasks, e.g. fingerprint matching or mosaicking \cite{cui2020dense}, and thus cannot directly provide interpretable pixel-wise correspondences of feature points in the matching result based on the model's output, such as fixed-length representations. Grosz et al. \cite{grosz2023afrnet} devise AFRNet to extract discriminative fingerprint representations for fingerprint indexing, enabling visualization and refinement using correspondences between local fixed-length fingerprint representations in low-certainty situations. However, the convolutional neural network branch of AFRNet limits the number of obtainable correspondences due to its local focus. Additionally, the correspondences produced by AFRNet are patch-based rather than pixel-wise and are derived through a computationally intensive brute-force algorithm. Given these considerations, it is imperative to develop a reliable approach that can learn fixed-length representations and provide interpretable dense pixel-wise correspondences of feature points simultaneously, especially in low-texture areas of fingerprints.

Existing architectures for extracting fixed-length fingerprint representation are primarily based on Convolutional Neural Networks (CNNs), often accompanied by domain knowledge (e.g. minutiae information) \cite{engelsma2019learning, lin2018cnnframework, takahashi2020fingerprint} to improve the matching performance. Other than CNNs, the application of Vision Transformer (ViT) has also emerged recently as an effective solution \cite{tandon2022transformer, grosz2023afrnet}, owing to  its superior performance in comparison to CNNs in various computer vision tasks \cite{dosoViTskiy2020image, chen2022mmViT}. Unlike CNNs focusing on local regions, the ViT utilizes a self-attention mechanism to attend to various parts of an input image, enabling it to capture long-range dependencies and global context \cite{huang2023model, yang2021focal}. Such characteristics may be particularly useful for extracting features pertaining to long-range texture information, e.g. ridges and valleys of fingerprints.

\begin{figure*}  
  \begin{center}
  \includegraphics[scale=0.54]{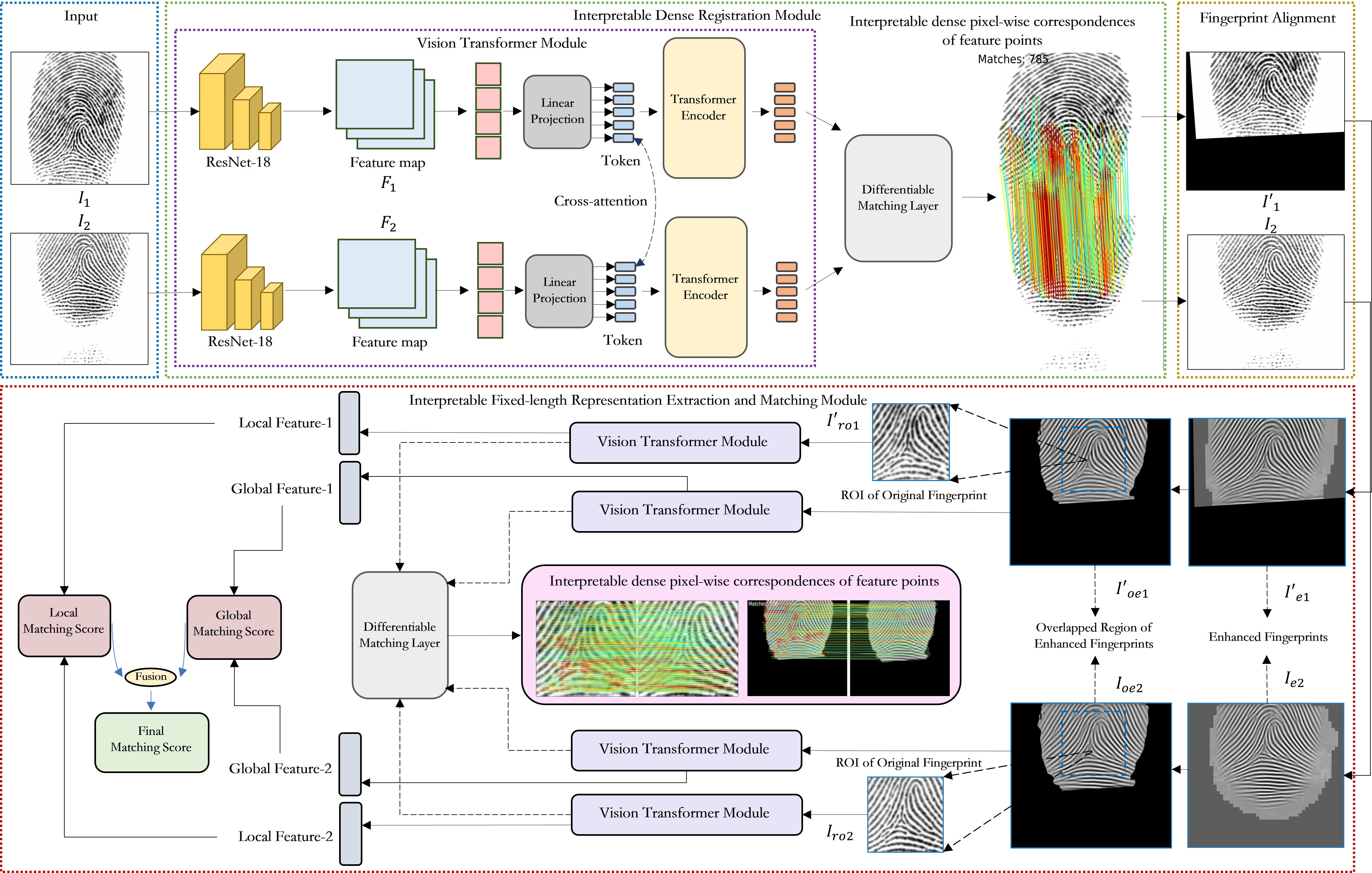}
  \caption{Overview of the IFViT architecture. The input fingerprint pair is processed by the ViT to obtain interpretable dense pixel-wise correspondences of feature points, which are in turn used to align the fingerprint pair. The aligned fingerprint pair is then enhanced by FingerNet and passed into the ViT by taking into account both local and global representations to obtain the discriminative fixed-length representation and interpretable dense pixel-wise correspondences of feature points in the matching result.}
  \vspace{-0.7cm} 
  \label{fig:IFViT}
  \end{center}
\end{figure*}

To the best of our knowledge, there is no deep learning-based fingerprint matching method that could simultaneously produce interpretable dense pixel-wise correspondences of feature points and discriminative fixed-length representations directly for the given fingerprint pair. Instead of primarily aiming to improve recognition performance like previous studies \cite{cao2017fingerprint,engelsma2019learning,grosz2023afrnet}, we focus more on improving the interpretability of fingerprint matching whilst maintaining reasonable recognition performance. Inspired by the global attention mechanism of ViT, we propose a multi-stage fingerprint matching network, namely Interpretable Fixed-length Representation for Fingerprint Matching via Vision Transformer (IFViT), to learn interpretable fixed-length representation to capture long-range relationships between different parts of the fingerprint. Fig. \ref{fig:IFViT} presents the entire architecture of IFViT. It is composed of two modules, namely an interpretable fingerprint dense registration module and an interpretable fixed-length representation extraction and matching module. In contrast to the previous fingerprint dense registration approaches implemented by CNN \cite{cui2020dense, gu2020latent}, this study is the first to utilize the ViT with the global attention mechanism for producing dense pixel-wise correspondences of feature points. In the challenging examples shown in Fig. \ref{fig:Matching Imgs}, a large number of correspondences of feature points could be still produced in the case of low-quality and cross-sensor fingerprint pairs based on the proposed method. Additionally, we also develop a ViT-based Siamese Network as the primary backbone for fingerprint matching, taking into account both local and global representations of fingerprints to learn fixed-length representations and provide interpretable pixel-wise correspondences of feature points in matching results.

We evaluate our proposed model on several public fingerprint datasets against previous deep learning-based fingerprint recognition methods. Experimental results demonstrate its effectiveness in terms of registration and matching performances.  Moreover, the proposed method could offer interpretable dense pixel-wise correspondences of feature points in the alignment and matching results, a distinctive feature absent in previous works on learning fingerprint fixed-length representations. We summarise our contributions as follows:

\begin{itemize}
    \item We introduce a ViT-based interpretable multi-stage deep learning framework for fingerprint matching. It could not only learn the discriminative fixed-length representation but also provide interpretable dense pixel-wise correspondences of feature points on the given fingerprint pair in matching results.  
    \item We propose an interpretable Siamese network-based fingerprint dense registration module via ViT, which can provide interpretable dense pixel-wise correspondences of feature points in the fingerprint pair, even for low-texture areas. These learned correspondences can be further employed for effective fingerprint alignment and improving interpretability in the fingerprint matching stage.
    \item We propose an interpretable Siamese network-based fixed-length representation extraction and matching module via ViT. It combines local and global representations from fingerprints for semantic shortcomings of each other. The matching performance has been evaluated on diverse public databases and is comparable to state-of-the-art fingerprint recognition methods based on deep learning,  when the model is trained on a limited number of fingers.
\end{itemize}






\section{Related Work}

\subsection {Fixed-Length Representations of Fingerprints}

A number of studies \cite{engelsma2019learning,grosz2021c2cl,cao2019endtoend,takahashi2020fingerprint,grosz2022minutiae} have investigated the feasibility of  learning deep fixed-length fingerprint representations in fingerprint recognition. CNN-based approaches are among the first that were widely employed  \cite{cao2017fingerprint, engelsma2019learning, grosz2021c2cl, lin2018cnnframework}. As the pioneering work, Cao et al. \cite{cao2017fingerprint} design an Inception V3-based approach for fingerprint indexing and matching using the extracted fixed-length representation \cite{cao2017fingerprint}. To improve the recognition performance, the domain knowledge e.g. minutiae information is further introduced in deep networks. One of the representative works is DeepPrint \cite{engelsma2019learning}, which could achieve competitive performance compared to the state-of-art minutiae matchers. The learning of fixed-length representation is also taken into account in contactless and contact fingerprint matching \cite{grosz2021c2cl}. Different from the aforementioned approaches that are based on classification, similarity metric learning has been also used to learn fixed-length representations. Lin et al. develop a Siamese CNN-based framework for accurately matching contactless and contact-based fingerprint images \cite{lin2018cnnframework}.

More recently, the application of the ViT for learning fixed-length fingerprint representation has also drawn attention \cite{grosz2022minutiae, tandon2022transformer}. Grosz et al. \cite{grosz2022minutiae} propose the first use of the ViT in extracting fixed-length fingerprint representation and demonstrate that the independent performance of ViT in fingerprint recognition rivals that of CNN-based models. Tandon et al. \cite{tandon2022transformer} propose a convolutional ViT network based on both global representation (i.e. complete fingerprint image) and local representation (i.e. minutiae information) to complement semantic information in fingerprint recognition. 

However, most of above-mentioned approaches encounter the limitation of interpretability during the fingerprint matching process. The difficulty lies in the lack of ability of these learned fixed-length representations to determine feature points of the fingerprint pair utilized for matching. They lack the same level of interpretability exhibited by the minutiae-based matching method as the latter can visually provide pixel-wise minutiae correspondences in the matching result. 

\subsection {Interpretability of Fingerprint Matching}
Establishing deep learning models with acceptable interpretability has become an important task for numerous studies when dealing with computer vision-related tasks in recent years \cite{confalonieri2021historical, samek2019towards}. As another prevalent biometric technology for identifying individuals, facial recognition has received significant attention, leading to considerable research on interpretable deep learning models for face matching and indexing  \cite{neto2023pic-score, yin2019towards}. In contrast, there is hardly any research on interpretable fingerprint recognition technology. The need for interpretable fingerprint recognition approaches based on deep learning models is equally significant and warrants more consideration. 

The minutiae-based fingerprint matching approaches have held a dominant position over the past few decades \cite{jiang2000fingerprint, cappelli2010minutia}. This is attributed not only to its reliable performance in matching but also to its remarkable interpretability. The minutiae expressed by their origin (i.e. $x$ and $y$ coordinates) and angle are extracted first and then minutiae correspondences in the input fingerprint pair could be established and visualized. Similar to minutiae matching, Cao et al. \cite{cao2019endtoend} introduce the concept of virtual minutiae using deep learning models to sample directed points located on fingerprint images to ensure an adequate number of key points within the latent fingerprint area and to provide interpretable correspondences. Gu et al. \cite{gu2020latent} propose a coarse-to-fine matching scheme considering using undirected sampling points as key points to produce interpretable correspondences on the given latent fingerprint pair. However, while identified key points e.g. minutiae or virtual minutiae could be used to obtain the average similarity employed for fingerprint matching, they lack the benefit of the fixed-length representation extracted by deep models. Performing pairwise comparisons of fingerprints is computationally demanding which varies with the number of detected key points. On the other hand, Cui et al. \cite{cui2019dense} also define the task of fingerprint dense registration for obtaining pixel-wise displacement measures between two fingerprints with nonlinear skin distortion. The learned correspondences could contribute to the improvement of the subsequent matching performance as well as providing interpretable pixel-wise correspondences of feature points. 

Despite the aforementioned studies being able to provide interpretable pixel-wise correspondences of feature points, none of them is integrated with learning fixed-length fingerprint representation during the matching process, which limits the reduction of computational complexity and applicability in other downstream tasks e.g. encryption \cite{engelsma2019learning}. In addition, the CNNs employed in these studies have limited receptive fields, which is not conducive to finding correspondences of feature points within a large global context \cite{sun2021loftr}. A network called AFRNet is devised by Grosz et al. \cite{grosz2023afrnet} for learning discriminative fixed-length fingerprint representations in the indexing task while obtaining correspondences between local fixed-length fingerprint representations that can be used for refining global representations in low-certainty situations. However, similar to previous studies, the CNN branch of AFRNet constrains the achievable number of correspondences, limiting its interpretability. On the other hand, the correspondences are patch-based instead of pixel-wise and are not a direct output of the neural network. They are obtained using a brute-force algorithm that demands significant computational resources, making the process time-consuming. 

Employing gradients of the prediction output to identify parts of an input image that exert the most influence on the fingerprint matching result is also considered. Chowdhury et al. \cite{chowdhury2020can} train a multi-scale dilated Siamese CNN using fingerprint patches and demonstrate the significance of minutiae for fingerprint matching using Grad-CAM \cite{selvaraju2017gradcam}. However, it could only reveal general areas that CNN focuses on during the inference stage, but could not establish further detailed pixel-wise correspondences.

In contrast to previous studies, we propose IFViT for fingerprint matching with excellent matching performance and interpretability. It combines the benefit of fixed-length representations extracted by deep models and interpretable pixel-wise correspondences of feature points in matching results. The ViT-based model is established to obtain interpretable dense pixel-wise matches. These matches could be employed for effective fingerprint alignment and improve interpretability in the subsequent matching procedure, allowing feature points used for fixed-length representation-based fingerprint matching to be visualized.

\section{Interpretable Fixed-Length Representation for Fingerprint Matching}
In this section, we provide a detailed description of the proposed IFViT. Firstly, we provide a high-level overview and intuition of IFViT, then discuss the implementation mechanism of the ViT-based module for interpretable fingerprint dense registration, followed by the working mechanism of the ViT-based module for interpretable fingerprint fixed-length representation extraction and matching.
\vspace{-0.3cm} 
\subsection {Overview}
The architecture of the proposed IFViT is shown in Fig. \ref{fig:IFViT}. It contains two main modules, an interpretable dense registration module and an interpretable fixed-length representation extraction and matching module. IFViT is trained with a large dataset composed of millions of fingerprint images across several benchmark datasets including plain, slap and rolled fingerprints.

Given the input fingerprint pair $I_1$ and $I_2$, a specific CNN, ResNet-18 \cite{he2016deep}, is employed to extract local features $F_{1}$ and $F_{2}$. The ViT is then adopted to learn the position and context-dependent features from $F_{1}$ and $F_{2}$, followed by a differentiable matching layer \cite{tyszkiewicz2020disk} to produce dense pixel-wise correspondences of feature points for the given fingerprint pair. Alignment is subsequently carried out based on these correspondences to obtain an aligned fingerprint image $I'_{1}$ with reference to the other input $I_{2}$.  The aligned fingerprint pair then undergoes fingerprint enhancement implemented by FingerNet \cite{tang2017fingernet}, a state-of-the-art minutiae extractor consisting of orientation estimation, segmentation, Gabor enhancement and minutiae extraction, to obtain the enhanced fingerprint pair $I'_{e1}$ and $I_{e2}$. To take full advantage of the complementary local and global representation of the input fingerprint pair, we identify the corresponding Region of Interests (ROIs), including $I'_{ro1}$ and $I_{ro2}$ in original fingerprint images, and the overlapped regions  $I'_{oe1}$ and $I_{oe2}$ in the aligned enhanced fingerprint images. Subsequently, those ROIs are passed into the ViT that is trained in the interpretable dense registration module but is finetuned with the new fully connected layer to obtain interpretable dense pixel-wise correspondences of feature points and discriminative fixed-length fingerprint representations, simultaneously. The learned local and global fixed-length representations are individually computed for their similarity, and combined to derive the final matching score.

\vspace{-0.4cm}
\subsection {Interpretable Dense Registration} 

Dense fingerprint registration seeks to establish pixel-wise correspondences of feature points between two fingerprints that are subjected to nonlinear skin distortion or displacement \cite{si2017dense}. These learned correspondences could then be used to fit a spatial transformation model for aligning fingerprints which is an essential component in most fingerprint recognition systems \cite{engelsma2019learning}.  Since the quality of the alignment may greatly affect the performance of subsequent fingerprint matching process, it motivates us to develop a robust dense registration model. We particularly consider the characteristics of the ViT due to its ability to capture global context.

Given the pair of fingerprint images $I_{1}$ and $I_{2}$, firstly we use a CNN to learn translation-equivariant local features which also reduces the input length to the ViT module to ensure a manageable computation cost. In this study, the ResNet-18 is chosen as the backbone. After obtaining feature maps $F_{1}, F_{2} \in \mathbb{R}^{H \times W \times D} $ from the CNN, a ViT model is built to extract local features that are both position-dependent and contextually relevant. Specifically, ViT performs a linear projection at first based on $N$ patches $f_i \in \mathbb{R}^{1 \times 1}$ in feature maps $F_{1}, F_{2} $ and then transforms them into 1D tokens ${z_i \in \mathbb{R}^d}$ as shown below.
\begin{equation}
\textbf{z}=[Ef_1,Ef_2,...,Ef_N] + p
\end{equation}
where $E$ represents the linear projection. The position embedding $p \in \mathbb{R}^{N \times d}$ is also added to tokens to provide the positional information. Subsequently, the token is passed into the encoder consisting of a sequence of $L$ transformer layers. As a critical part in the encoder, the attention layer receives the token in three different vectors: query ($q$), key ($k$) and value ($v$). The vectors originating from diverse inputs are subsequently consolidated into three different matrices, namely $Q$, $K$ and $V$. The query $Q$ retrieves information from the value $V$ based on the attention weights calculated from the dot product of $Q$ (corresponding to each value $V$) and the key vector $K$. The process is denoted as:
\vspace{-0.1cm}
\begin{equation}
Attention( Q, K, V) = softmax( Q K^T) V
\end{equation}

In addition to the self-attention layer, the cross-attention layer inspired by \cite{sarlin2020superglue, sun2021loftr} is also introduced in our ViT model to enable cross-image communication in the fingerprint pair similar to the way humans look back-and-forth when matching images. In terms of the self-attention layer, the input features $f_i$ and $f_j$ are the same (either $F_{1}$ or $F_{2}$). However, for the cross-attention layer, the  $f_i$ and $f_j$ are either ($F_{1}$ and $F_{2}$) or ($F_{2}$ and $F_{1}$). The self-attention layer and cross-attention layer are interleaved in our transformer by $M$ times. After passing through the encoder layer, the transformed features denoted by $F_{1}'$ and $F_{2}'$ are obtained.

In terms of identifying the dense pixel-wise correspondences of feature points in the given fingerprint pair, a differentiable matching layer using the dual-softmax operator is adopted \cite{tyszkiewicz2020disk}. Firstly, we compare the feature similarity for each pixel in $F_{1}'$ with respect to all pixels in $F_{2}'$ by calculating their correlations as follows:

\begin{equation}
C= \frac{{F}_{{1}}' {{F'}_{2}^T}}{\sqrt{D}} \in \mathbb{R}^{(H \times W) \times (H \times W)}
\end{equation}
where correlation matrix $C$ denotes the correlation value of each coordinate between $F_{1}'$ and $F_{2}'$. $\frac{1}{\sqrt{D}}$ is a normalization factor to prevent the dot-product operation from producing large values.

Subsequently, we normalize $C$ using the softmax operation to obtain the matching confidence matrix $P$ as in equation (\ref{eq:matching_confidence}). The utilization of a softmax-based approach not only enables the end-to-end training process but also provides pixel-level accuracy. 

\begin{equation}\label{eq:matching_confidence}
P=softmax(C) \in \mathbb{R}^{(H \times W) \times (H \times W)}
\end{equation}

The candidate matches in $P$ are selected if the corresponding confidence score is higher than a given threshold $T_1$. We use Mutual-Nearest-Neighbor (MNN) criteria to further filter outlier matches and obtain the final selected matches $M$:
\vspace{-0.05cm}
\begin{equation}
M= \{ (i,j)| \forall (i,j) \in \mathrm{MNN} (P), P(i,j) \geqslant T_1 \}
\end{equation}

\begin{figure}
  \begin{center}
  \includegraphics[width=3.5in]{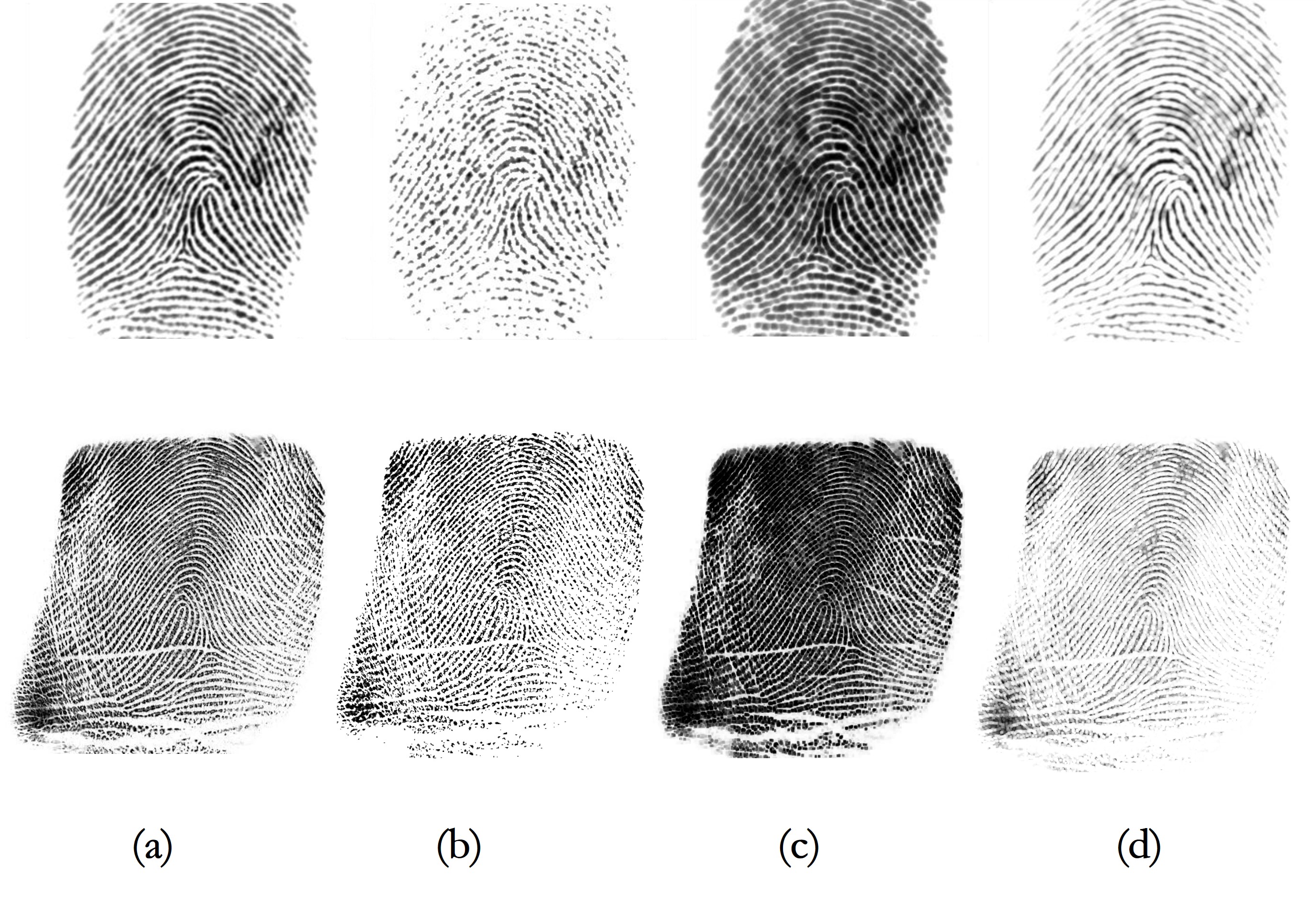}
  \caption{{Examples of synthetic corrupted fingerprints simulated by diverse types of noises. (a) Original fingerprint (b) fingerprint processed by sensor noise (c) fingerprint processed by over-pressurization operation (d) fingerprint processed by dryness operation.}}
  \vspace{-0.7cm}
  \label{fig:noiseimage}
  \end{center}
\end{figure}

$\textbf{Loss function}$. For the interpretable dense registration module, the loss function for training the ViT that produces the dense pixel matches is negative log-likelihood loss, which is computed based on the matching confidence matrix $P$. The pre-trained ViT model for feature matching in LoFTR \cite{sun2021loftr} is employed to accelerate the convergence for training. Considering there is no available public fingerprint database that provides ground truth labels of matches of fingerprint pairs, we construct several noise models to simulate diverse types of noises and corruptions appearing in the real fingerprint images to improve the model's ability for recognizing correspondences of feature points in low-quality or low-texture fingerprint images. In this case, the fingerprint images enhanced by FingerNet are not used but instead, we solely rely on original fingerprint images and their corresponding corrupted image to train the model. Since the relation of each pixel position between original and synthetic corrupted fingerprint images could be directly obtained, we are able to determine ground-truth labels $G^{gt}$ for the confidence matrix. Three types of noises are simulated in this study including sensor noise, dryness and over-pressurization, as depicted in Fig. \ref{fig:noiseimage}. Sensor noise could be approximated by the Perlin noise \cite{cappelli2004improved}, whilst dryness and over-pressurization could be simulated by performing the dilation and erosion operation, respectively. To ensure rotation invariance, all corrupted fingerprint images are rotated randomly within $ \pm$60 degrees. This rotation parameter is determined based on the domain knowledge regarding the maximum degree to which a user would rotate their fingers while positioning them on the reader platen \cite{engelsma2019learning}. Furthermore, to minimize incorrect matches, ground-truth labels are also created for the registration of imposter fingerprint pairs. The loss function of dense registration $\mathcal{L}_D$ based on negative log-likelihood loss over the grids in $G^{gt}$ is minimized during the training stage:

\begin{equation}
\mathcal{L}_D = -\frac{1}{|G^{gt}|}\sum_{(i,j) \in G^{gt}} log {P(i,j)}
\end{equation}

\vspace{-0.2cm}
\subsection {Interpretable Fixed-length Representation Extraction and Matching}
After performing the dense registration on the given fingerprint pair, we establish the second critical component of our method, namely the interpretable fixed-length-representation extraction-and-matching module. In this module, we combine both local and global representations of fingerprints to improve the matching performance. It can additionally provide interpretable dense pixel-wise correspondences of feature points similar to the previous interpretable dense registration model in the matching results.

$\textbf{Fingerprint Alignment}$. Recognizing that the outcome of the alignment would significantly influence the subsequent matching module's performance, we rotate $I_{1}$ at five angles (-60°, -30°, 0°, 30°, 60°) to determine the best matches $M_{max}$ for alignment. Subsequently, we compute the homography matrix $H$ using the RANSAC algorithm and apply the obtained $H$ directly to $I_{1}$ to produce the aligned $I_{1}'$.
Compared with the Spatial Transformer Network (STN) which has been widely used recently for fingerprint alignment, e.g., in \cite{engelsma2019learning, takahashi2020fingerprint}, our proposed ViT-based interpretable dense registration module for fingerprint alignment can provide dense pixel-wise correspondences of feature points for the given fingerprint pair, which is conducive to understanding the criteria used to determine when fingerprint alignment succeeds or fails.

\begin{figure}
  \begin{center}
  \includegraphics[width=3.5in]{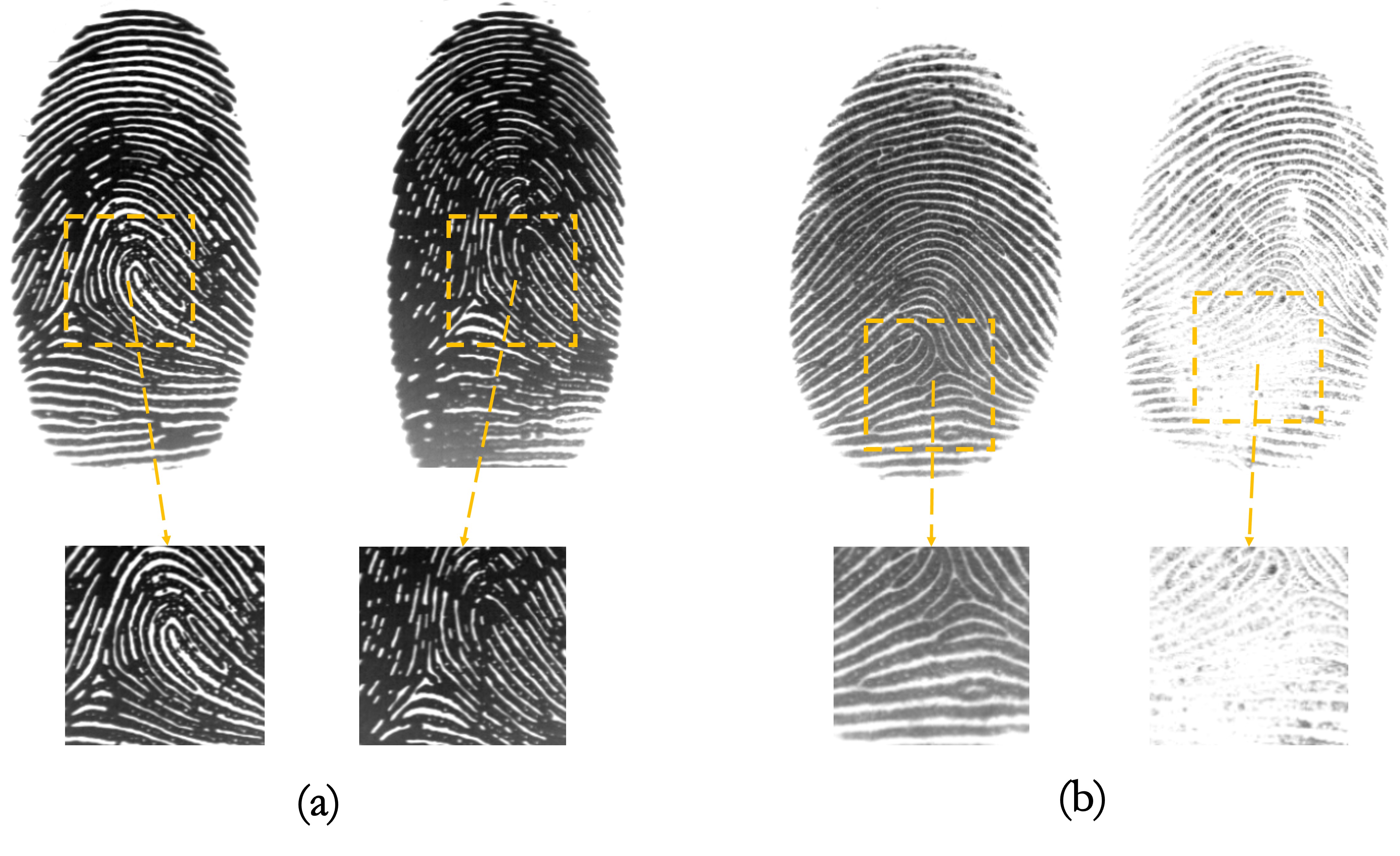}
  \caption{{Cases prone to matching failure: (a) The similar global ridge-flow structures of the fingerprint from different identities. (b) The dissimilar representations in the local patch of different impressions from the same finger.}}\label{fig:waveforms}
  \vspace{-0.7cm} 
  \end{center}
\end{figure}

$\textbf{Obtaining Local and Global Representation}$. Although the global fixed-length fingerprint representation has been demonstrated to be effective for fingerprint recognition \cite{engelsma2019learning, song2017fingerprint, engelsma2021infant-id}, it may produce high similarity on some global similar ridge-flow structures from different identities by mistake, therefore leading to wrong matches \cite{tandon2022transformer}. On the other hand, local representation, e.g. fingerprint patches, may also face matching failures, primarily because some factors like distortion and dryness could negatively impact the effective fingerprint information, especially for minutiae \cite{tandon2022transformer}. Fig. \ref{fig:waveforms} shows a case susceptible to matching failures when considering either only the local or global representation. Therefore, combinations of both representations should be considered to compensate for the semantic shortcomings of each other.

In this study, we fuse global representation extracted from overlapped regions of enhanced fingerprints with local representation derived from the corresponding ROI of original fingerprints. The motivation for considering the ROI from original fingerprints instead of enhanced fingerprints as the local representation is that we observed a certain level of prediction error when using FingerNet for enhancement. The FingerNet performs well in enhancing the global ridge-flow structures of fingerprints but is prone to introducing patch-like artifacts, resembling a mosaic, in low-quality local areas in fingerprints. This could potentially lead to the loss of some crucial information.

\begin{figure}
  \begin{center}
  \includegraphics[width=3.55in]{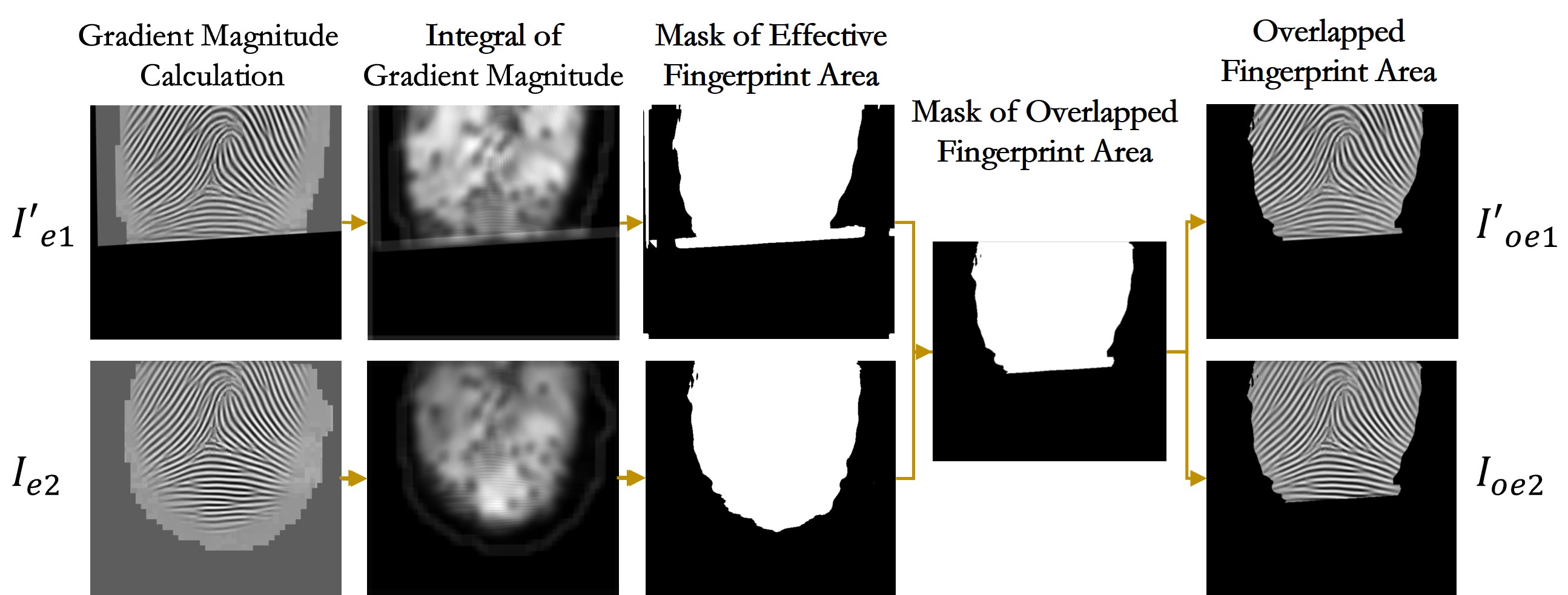}
  \caption{{The procedure of extracting the overlapped fingerprint areas based on the given fingerprint pair.}}\label{fig:overlap_fingerprint_extraction}
  \vspace{-0.7cm} 
  \end{center}
\end{figure}

Specifically, we introduce the Sobel operator to calculate the gradient of the fingerprint image's intensity in the horizontal ($G_x$) and vertical ($G_y$) directions, and then compute the magnitude of the gradient $|G|$ for each pixel based on equation (\ref{eq:image_gradient}). The box filter with 25×25 kernel size is used to obtain the integral of the gradient magnitude. Threshold $T_2$ which equals 15 percent of the maximum integral of the obtained gradient magnitude is further empirically determined for segmenting the general fingerprint area. We perform detection for contours on the segmented regions and conduct connected component analysis, resulting in the respective mask of an effective fingerprint area for $I_{e1}'$ and $I_{e2}$. After taking the intersection of masks, they are separately applied to $I_{e1}'$ and $I_{e2}$, producing effective overlapped  fingerprint areas $I_{oe1}'$ and $I_{oe2}$ (i.e. global representation). The flowchart for extracting overlapped regions from the fingerprint pair is illustrated in Fig. \ref{fig:overlap_fingerprint_extraction}. Although the FingerNet can also directly output the mask of a fingerprint, the generated mask frequently includes irrelevant background around the fingerprint, as shown in $I_{e1}'$ and $I_{e2}$ of Fig. \ref{fig:overlap_fingerprint_extraction}. For the model to focus only on the effective fingerprint information, the mask generated by FingerNet is not considered in this step. Furthermore, we extract ROIs (90×90) from the overlapped region of original fingerprints as the local representation.

\begin{equation}\label{eq:image_gradient}
|G|= \sqrt{{G_x}^2+{G_y}^2}
\end{equation}

$\textbf{Learning Interpretable Fixed-length Representation}$. For the interpretable fixed-length representation extraction and matching module, we propose two ViT-based Siamese Networks $f(\cdot)$ with shared weights which learns a similarity metric for the input fingerprint pair and outputs a matching score. Specifically, the input consists of the overlapped fingerprint $I_{oe1}'$ and $I_{oe2}$ (i.e. global information) as well as the corresponding ROIs  $I_{ro1}'$ and $I_{ro2}$ extracted from original fingerprint images (i.e. local information). We utilize the interpretable dense registration module trained previously in the fingerprint matching module to generate pseudo labels of fingerprint pairs for learning correspondences of feature points. We combine the previously trained ViT-based Siamese model as the backbone with a new fully connected layer, which is then retrained with labelled fingerprint pairs. Thus, it offers interpretable dense pixel-wise correspondences of feature points $M$ in matching results, and the 256-dimensional fixed-length representations $R_1$ and $R_2$.

$\textbf{Loss Function}$. In this module, dense registration loss $\mathcal{L}_D$ is employed again to learn dense pixel-wise correspondences of feature points for the fingerprint pair from ViT's output. The cosine embedding loss $\mathcal{L}_E$ shown in equation (\ref{eq:cosine_loss}) is chosen for training the model to transform identified correspondences of feature points into the 256-dimensional discriminative fixed-length representation.

\begin{equation}\label{eq:cosine_loss}
\begin{split}
\mathcal{L}_E(R_1,R_2,y_e) = -\frac{1}{N}\sum_{i=1}^{N} \left \{
\begin{array}{ll}
    1-cos(R_1^i,R_2^i), \quad if \; y_e^i = 1\\
    max(0,cos(R_1^i,R_2^i) -m), \\ \quad \quad \quad \quad \quad \quad \quad \quad if \; y_e^i = -1
\end{array}
\right.
\end{split}
\end{equation}

To minimize and maximize the distance of learned intra-class and inter-class representation separately, we further introduce the ArcFace loss $\mathcal{L}_A$ in equation (\ref{eq:arcface_loss}) to optimize the network via classification task, constituting the total loss $\mathcal{L}_T$ shown in equation (\ref{eq:total_loss}). 

\begin{equation}\label{eq:arcface_loss}
\begin{aligned}
\mathcal{L}_{\text{A}}(y_c) = -\frac{1}{N} \sum_{i=1}^{N} \log\left(\frac{e^{s\cos(\theta_{y_c^i} + m)}}{e^{s\cos(\theta_{y_c^i} + m)} + \sum_{k=1,k\neq y_c^i}^{K} e^{s\cos(\theta_k)}}\right)
\end{aligned}
\end{equation}

\begin{equation}\label{eq:total_loss}
\begin{aligned}
\mathcal{L}_{T}(R_1,R_2,y_e,y_c) = \lambda_{1}\mathcal{L}_D + \lambda_{2} \mathcal{L}_E(R_1,R_2,y_e) + \lambda_{3} \mathcal{L}_A(y_c) \\
\end{aligned}
\end{equation}
where $y_e \in \{ -1, 1 \}$ represents the ground truth of the cosine embedding loss $\mathcal{L}_E$ and $y_c$ is the true class label of the ArcFace loss $\mathcal{L}_A$, respectively. $N$ is the total number of the training samples while $K$ is the number of classes. $\theta_{y_c^i}$ denotes the cosine of the angle between the learned representation ($R_1$ or $R_2$) and the weight vector for the corresponding label $y_c$. In terms of  $\theta_{k}$, it calculates the cosine of the angle between the learned representation ($R_1$ or $R_2$) and the weight vector for all classes except the $y_c$. Additionally, the margin $m$ is set to 0.4 for both $\mathcal{L}_E$ and $\mathcal{L}_A$ in this study and the rescaling factor $s$ is set as 64. $\lambda_{1}$, $\lambda_{2}$ and  $\lambda_{3}$ in the $\mathcal{L}_T$ are empirically set as 0.5, 0.1 and 1 to balance the matching performance and interpretability, respectively.

Rather than introducing the domain knowledge (i.e. minutiae information) \cite{takahashi2020fingerprint, grosz2021c2cl, engelsma2019learning}, we hope that the fixed-length representation learned from the IFViT could utilize prior information related to dense pixel-wise correspondences of feature points in the fingerprint pair, to deal with the situation where it is challenging to learn minutiae information from low-quality fingerprints.

$\textbf{Matching Score Computation}$. For evaluation, the learned fixed-length global representation $(R_{1}^{g}$, $R_{2}^{g})$ and fixed-length local representation $(R_{1}^{l}$, $R_{2}^{l})$ of an input fingerprint pair can be extracted. The matching scores, namely $t_1$ and $t_2$ based on the $(R_{1}^{g}$, $R_{2}^{g})$ and $(R_{1}^{l}$, $R_{2}^{l})$ are calculated as shown in equation (\ref{eq:matching_score}) and normalized to the range [-1, 1]. The total matching score $S$ is finally obtained by weighted fusion of the individual scores ($t_1$ and $t_2$) as shown in equation (\ref{eq:final_matching_score}), where the weights $\alpha_1$ and $\alpha_2$ are determined by linear regression learning.

\begin{equation}\label{eq:matching_score}
\begin{aligned}
t_k = (R_{1}^i)^T \cdot R_{2}^i, \quad \quad k \in \{ 1,2 \}, i \in \{ g,l \}
\end{aligned}
\end{equation}

\begin{equation}\label{eq:final_matching_score}
\begin{aligned}
S = \sum_{k=1}^{2} \alpha_it_k
\end{aligned}
\end{equation}

\section {Experiments}
In this section, we first introduce several public fingerprint datasets employed in this study along with a detailed experimental setup. Then we conduct the performance evaluation of the interpretable dense registration module and the interpretable fixed-length representation extraction and matching module in the proposed IFViT.
\vspace{-0.3cm}
\subsection {Datasets}
$\textbf{Dense Registration Module}$. Regarding the training of the interpretable dense registration module, the FVC2002 (DB1, DB2 and DB3) \cite{fvc2002}, NIST SD301a (A, B, C, E, J, K M and N) \cite{fiumara2018nist}, a subset of NIST SD302a (A, B, C, D, E, F, U, V, L and M) \cite{fiumara2019nist} and a subset of MOLF (DB1 and DB2) \cite{sankaran2015multisensor} are used. The total number of fingerprint images in these datasets is 25, 090, and after applying three noise models previously proposed, a total of 100, 360 training images can be obtained. They are combined to form 100K fingerprint pairs (75K for genuine pairs and 25K for imposter pairs) for training.

$\textbf{Matching Module}$. In terms of the interpretable fixed-length representation extraction and matching module, we employ FVC2002 (DB1, DB2 and DB3) \cite{fvc2002}, NIST SD301a (A, B, C, E, J, K M and N) \cite{fiumara2018nist}, a subset of NIST SD302a (A, B, C, D, E, F, U, V, L and M) \cite{fiumara2019nist}, a subset of MOLF (DB1 and DB2) \cite{sankaran2015multisensor} and a recently released synthetic fingerprint dataset called PrintsGAN \cite{engelsma2022printsgan} as the training dataset and evaluate the model on FVC2004 (DB1, DB2 and DB3) \cite{fvc2004}, NIST SD4 \cite{Watson1992}, the remaining test partitions of NIST SD302a (last 200 fingers) and MOLF (last 200 fingers). There is no overlap between fingers selected for training and testing sets in NIST SD302a and MOLF. Furthermore, to strike a balance between model performance and computational demand, we only use the first impression from different fingers when creating imposter fingerprint pairs; otherwise, the size of the training dataset could even reach 100 M. Additionally, in PrintsGAN, only 2, 500 fingers are selected as a part of our training set and mainly used for model pre-training. In total, our aggregated training dataset contains 3.6M pairs (7.2M images). Table \ref{tab:datasets} provides the details of all databases and their usage in this study.

\begin{table*}[ht]
\centering
\caption{Information related to fingerprint datasets employed in this study}
\label{tab:datasets}
\begin{adjustbox}{width=\textwidth}
\begin{tabular}{ccccccc}
\toprule
Database & FVC2002 DB1A \cite{fvc2002} & FVC2002 DB2A \cite{fvc2002} & FVC2002 DB3A \cite{fvc2002} & FVC2004 DB1A \cite{fvc2004} & FVC2004 DB2A \cite{fvc2004} &FVC2004 DB3A \cite{fvc2004}\\
\midrule
Example Image & \includegraphics[width=0.10\textwidth]{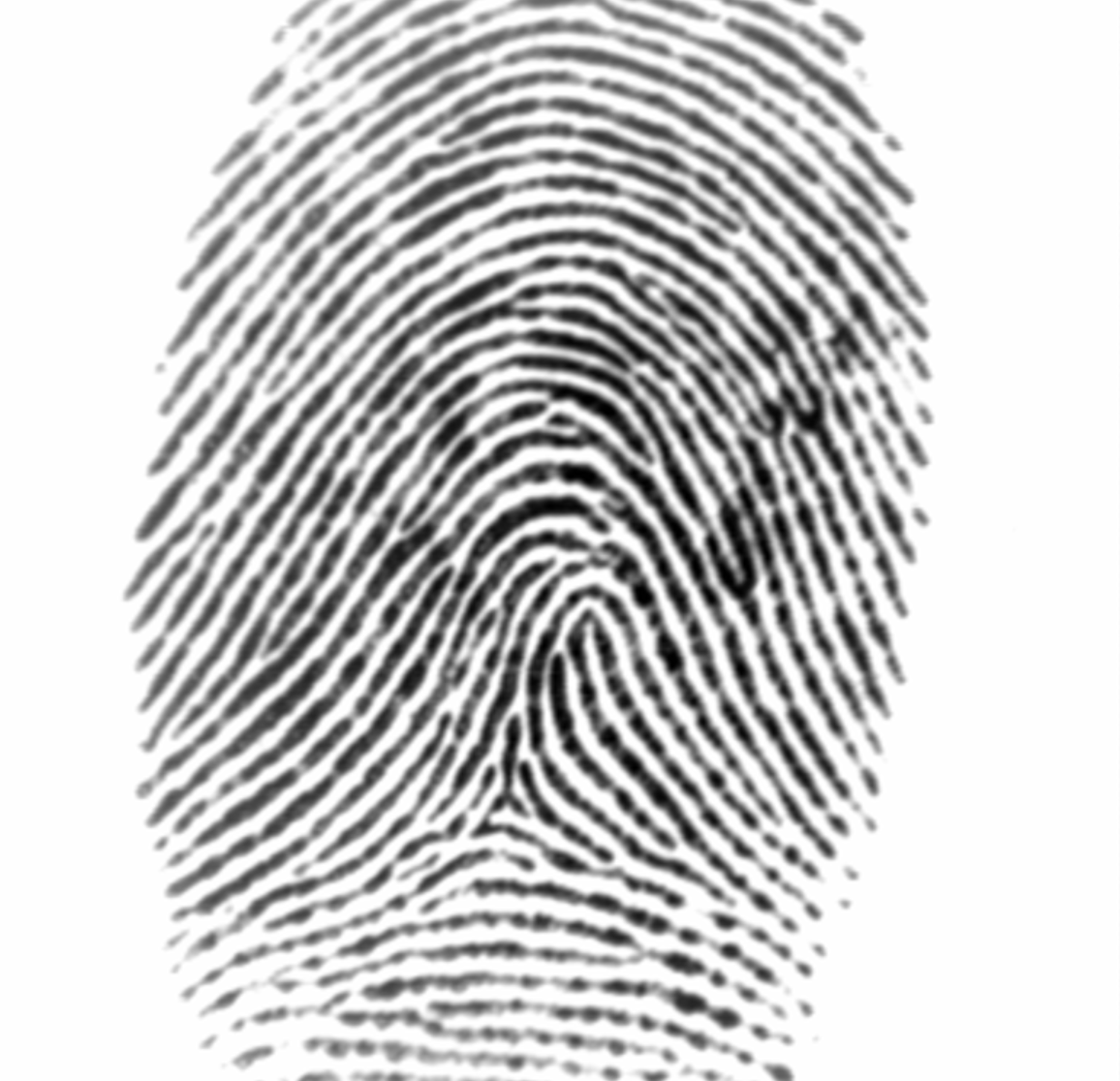} & \includegraphics[width=0.05\textwidth]{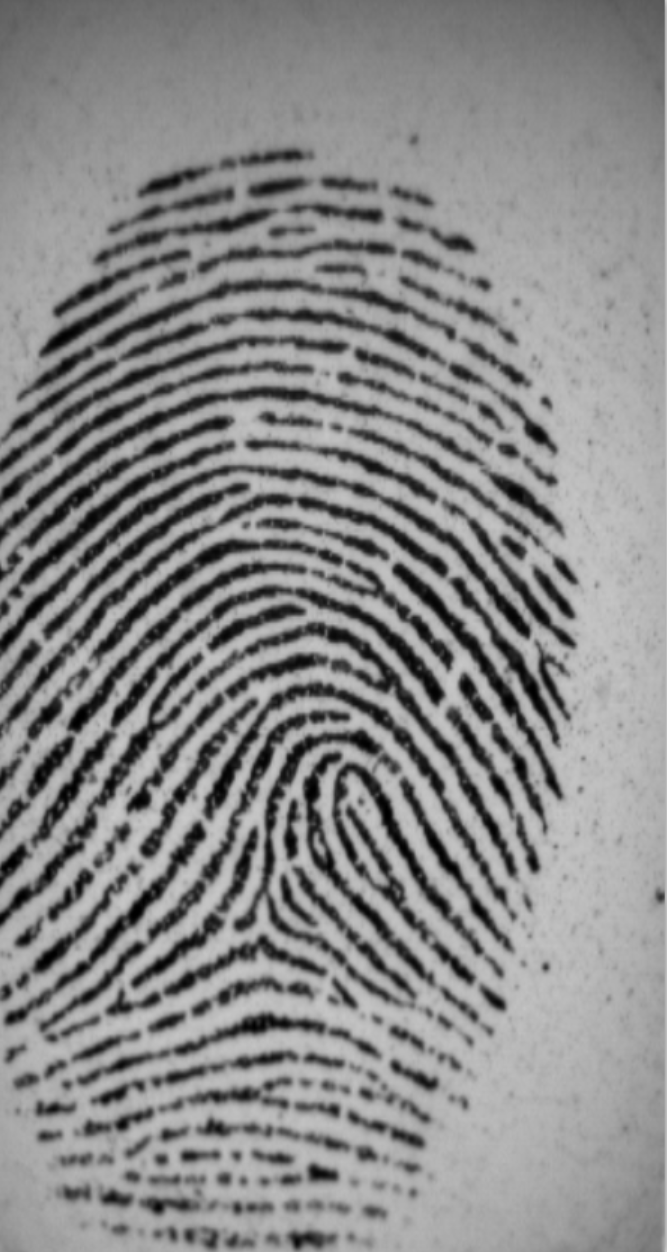}& \includegraphics[width=0.095\textwidth]{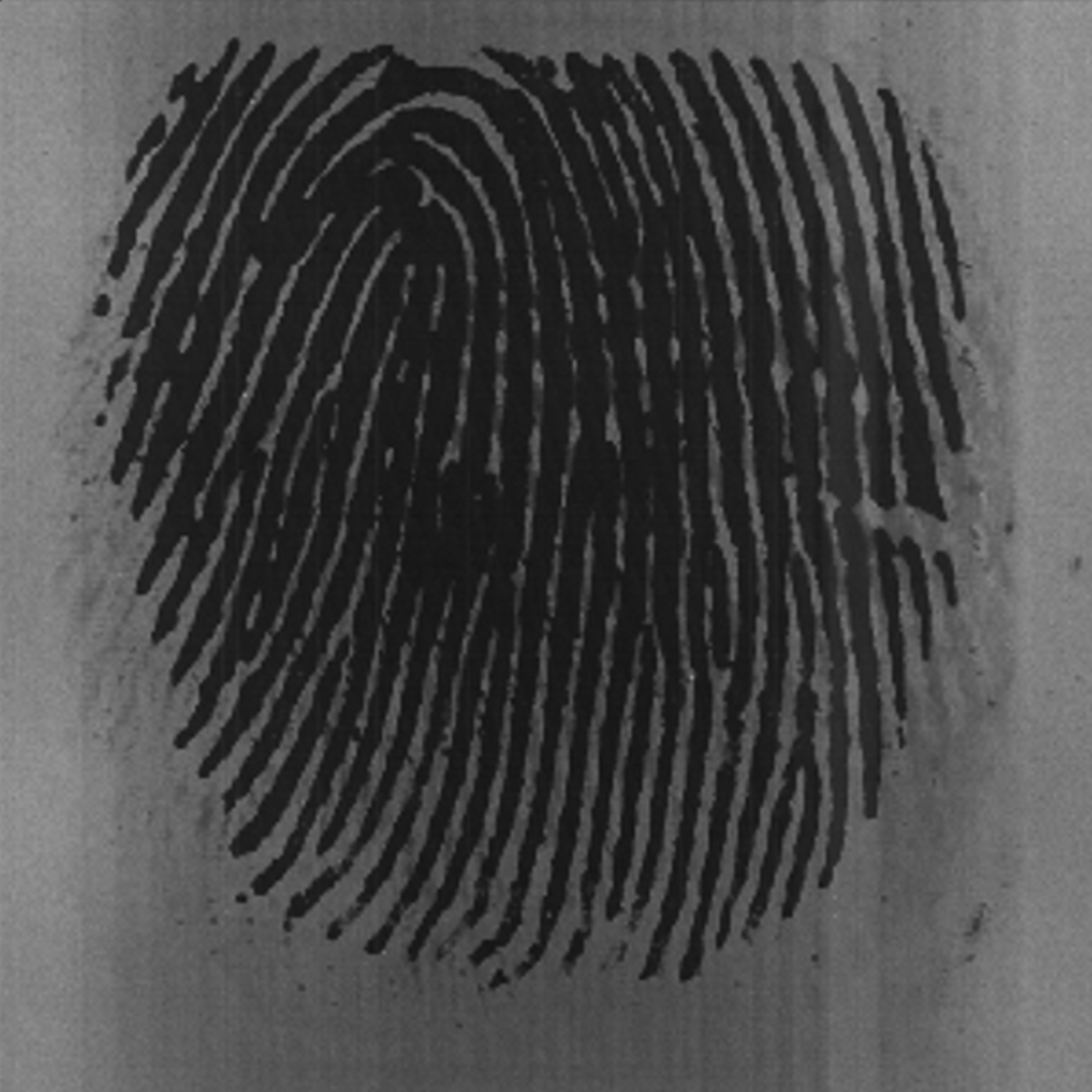}& \includegraphics[width=0.12\textwidth]{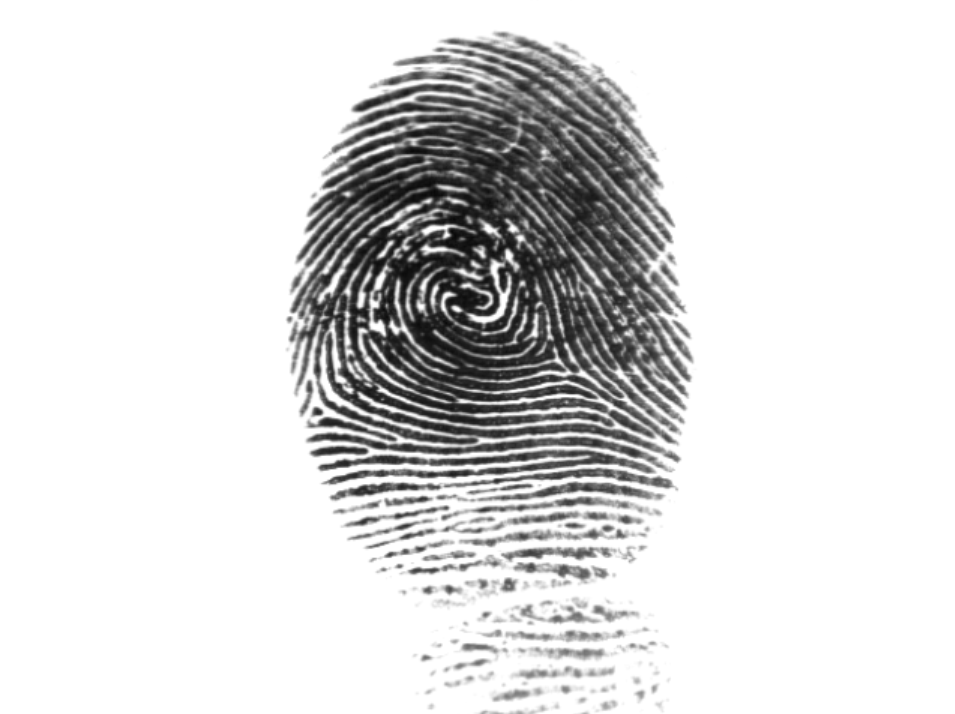}& \includegraphics[width=0.08\textwidth]{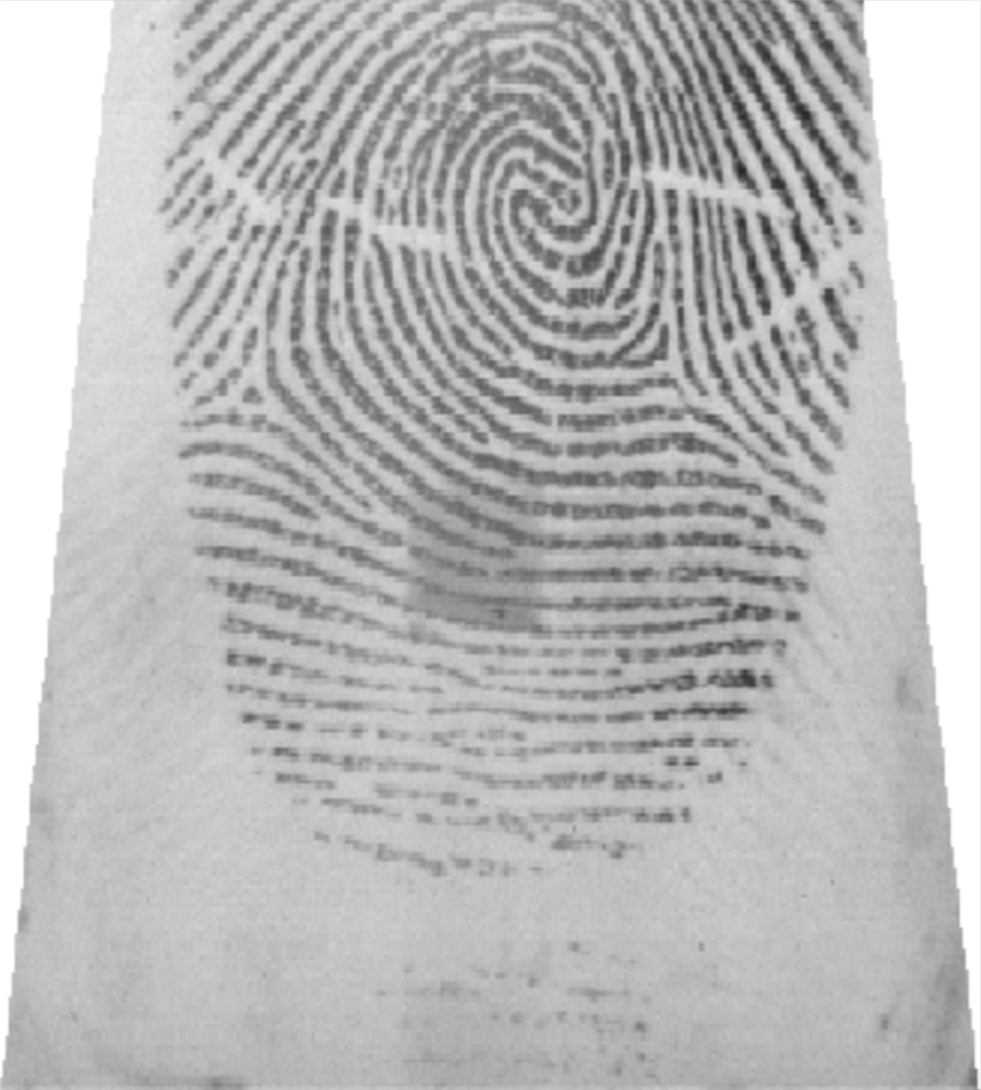} &
\includegraphics[width=0.057\textwidth]{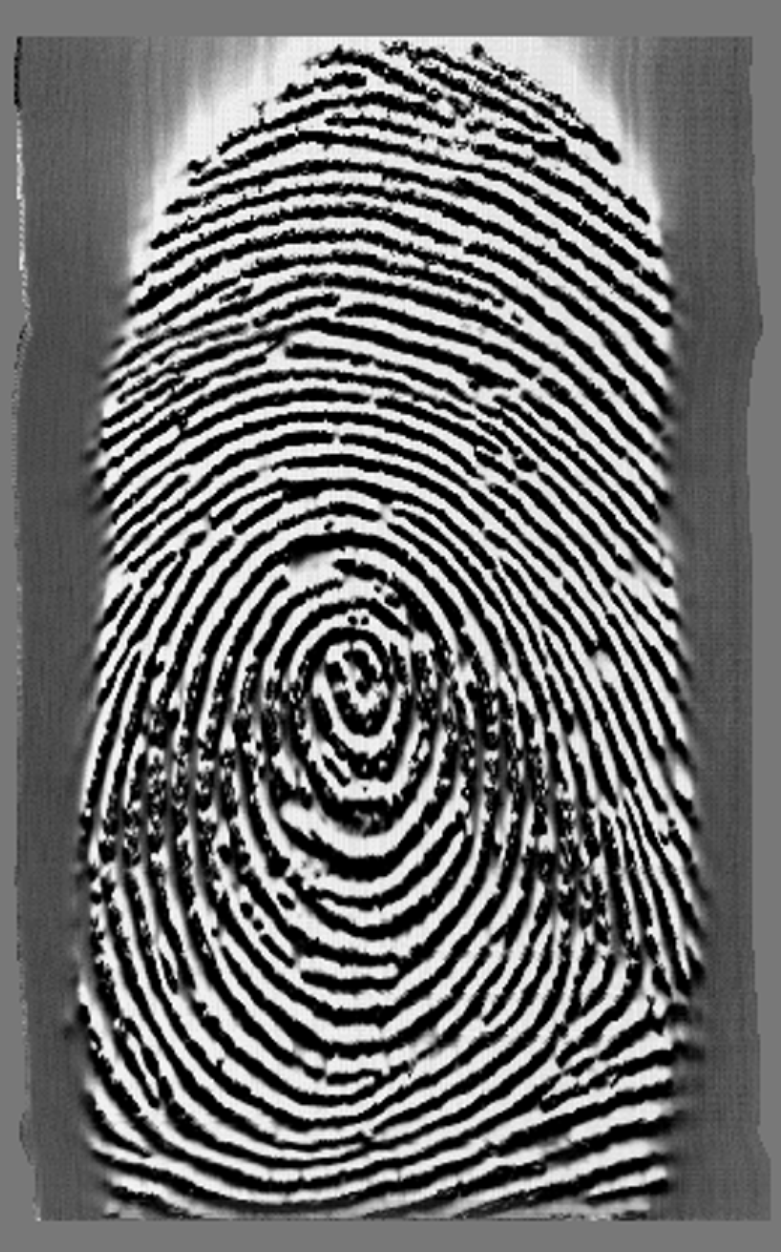}
\\
\midrule
Fingers Used& 100& 100& 100& 100&100&100\\
\midrule
Images Used& 800& 800& 800& 800&800&800\\
\midrule
Usage & Training (7, 750 pairs)& Training (7, 750 pairs)& Training (7, 750 pairs)& Testing (7, 750 pairs)&Testing (7, 750 pairs)&Testing (7, 750 pairs)\\
\toprule
\toprule
Database & NIST SD301a \cite{fiumara2018nist} & NIST SD302a \cite{fiumara2019nist}&  NIST SD4 \cite{Watson1992}& MOLF \cite{sankaran2015multisensor} &PrintsGAN \cite{engelsma2022printsgan}&\\
\midrule
 Example Image & \includegraphics[width=0.1\textwidth]{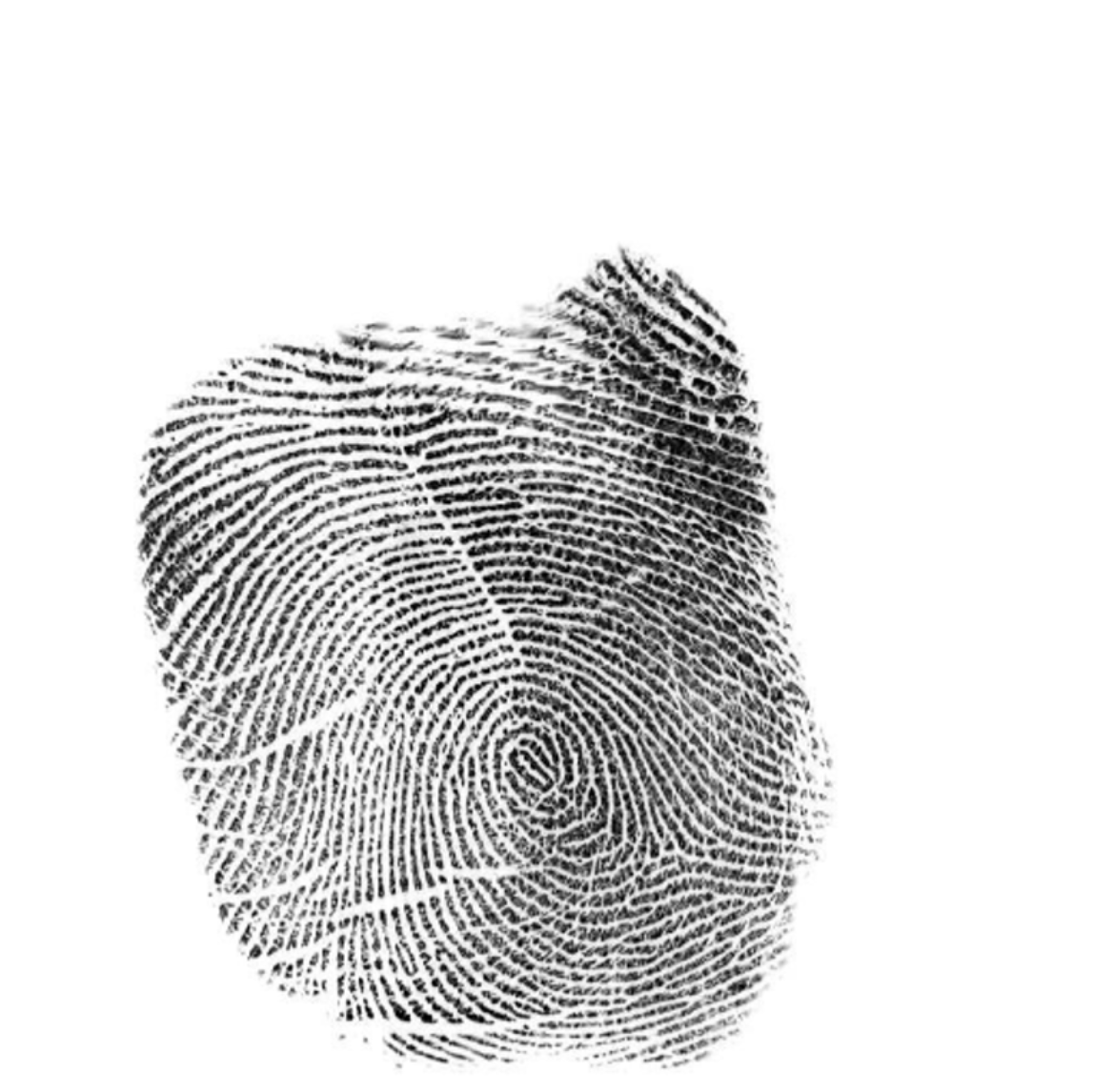}& \includegraphics[width=0.1\textwidth]{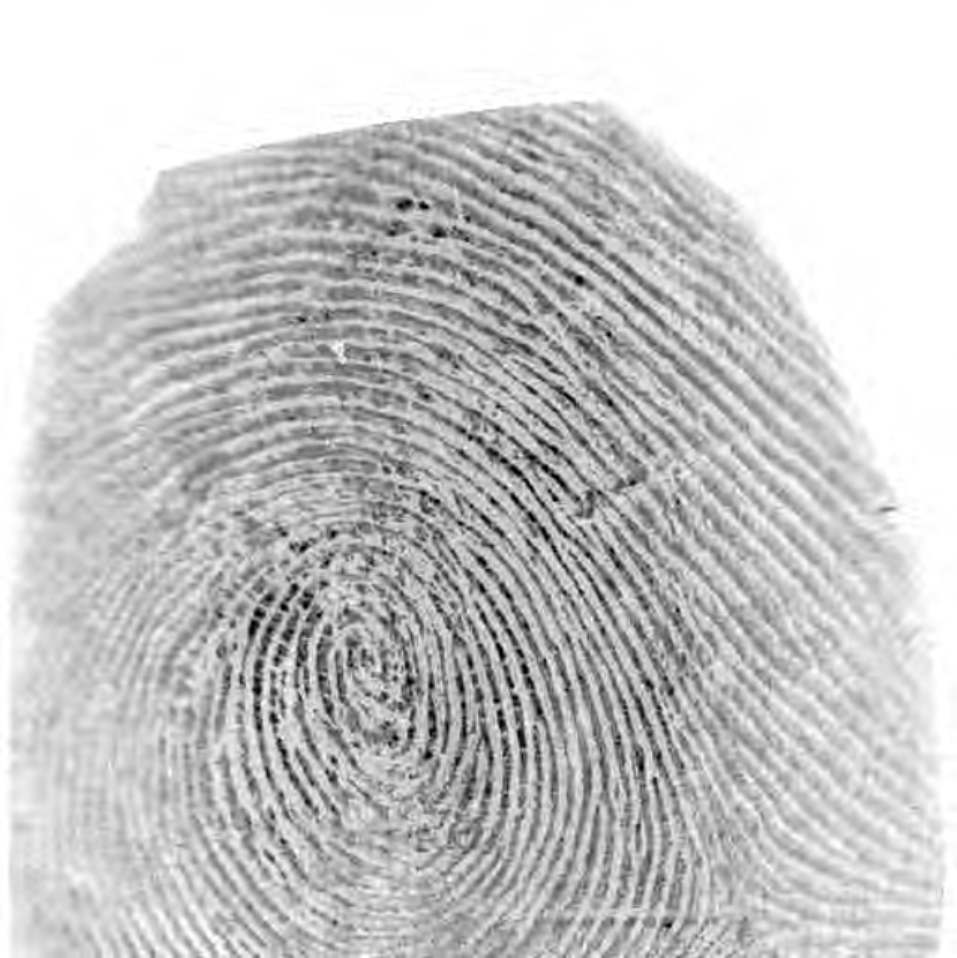}& \includegraphics[width=0.1\textwidth]{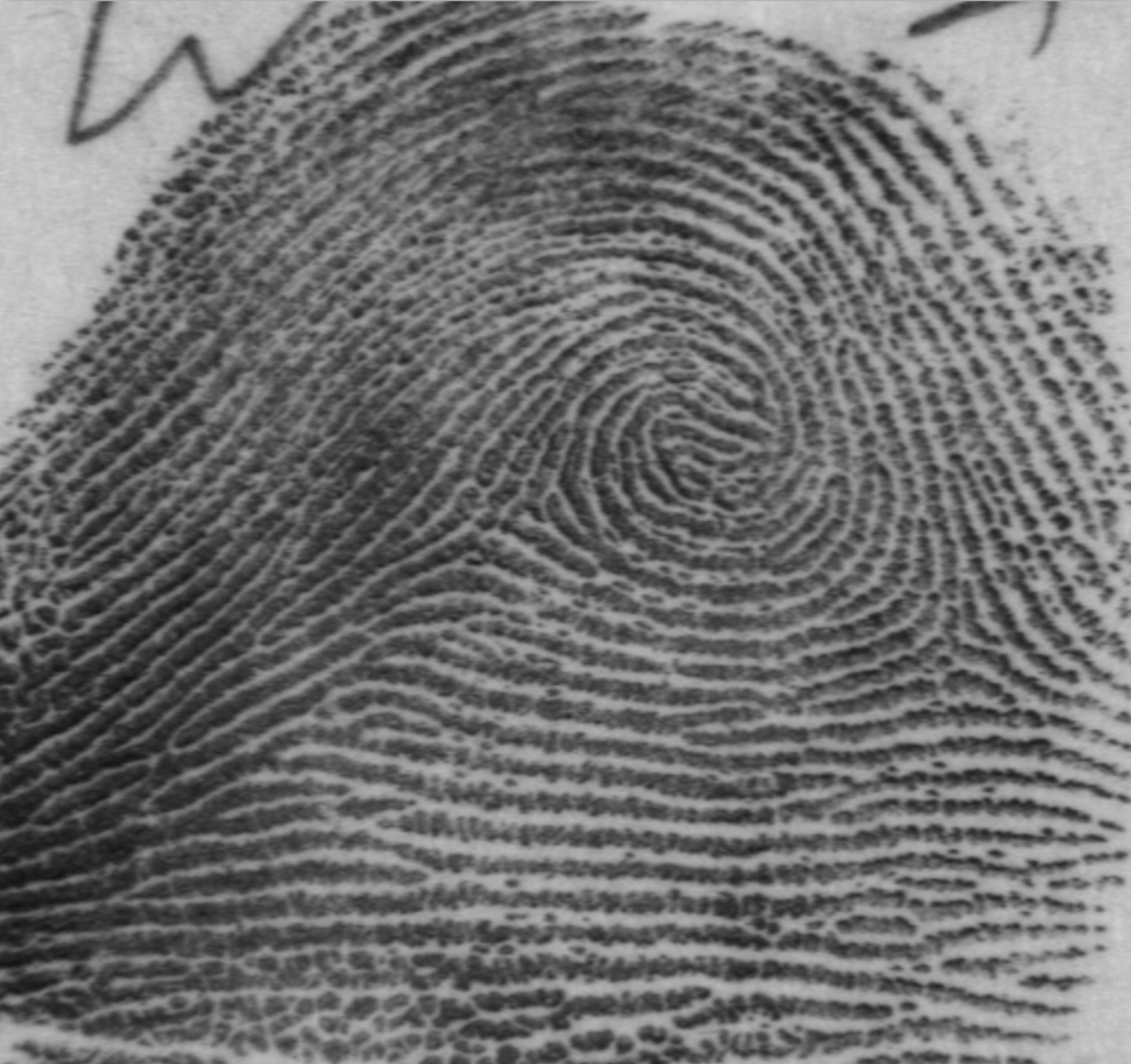}& \includegraphics[width=0.064\textwidth]{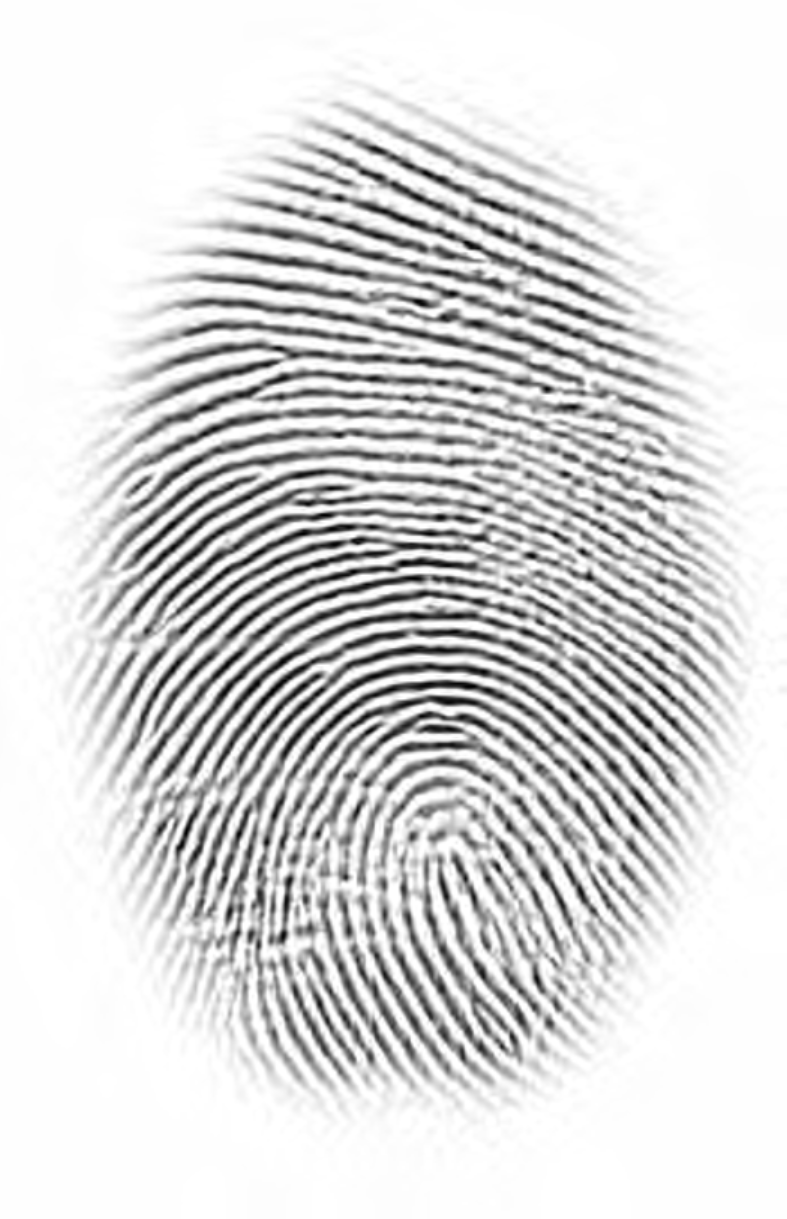}&\includegraphics[width=0.1\textwidth]{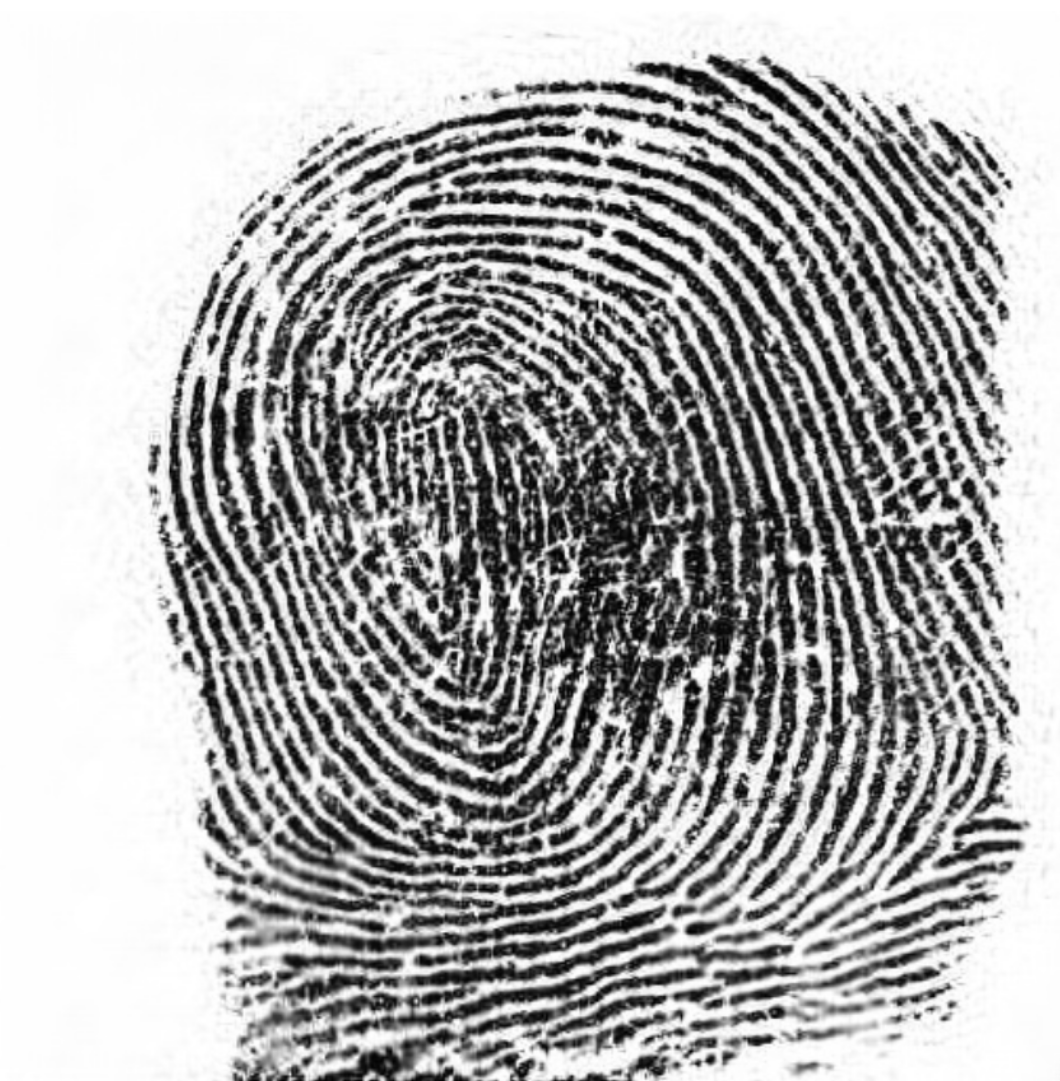} &\\
 \midrule
 Fingers Used& 240& 1, 789& 2, 000& 1, 000&2500&\\
 \midrule
 Images Used& 1, 920& 17, 890& 4, 000& 8, 000&37, 500&\\
 \midrule
 Usage & Training (9, 480 pairs)& \makecell{Training (193, 047 pairs) \\ Validation (18, 316 pairs) \\ Testing (28, 900 pairs)}& Testing (251, 000 pairs)& \makecell{ Training (62, 000 pairs) 
 \\ Testing (15, 500 pairs) }& Training (3, 386, 250 pairs)&\\
 \toprule
\end{tabular}
\label{table:dataset}
\end{adjustbox}
\end{table*}

\begin{table*}[ht]

\caption{Alignment performance evaluated using MI value on FVC2004 DB1, DB2, DB3, MOLF, NIST SD4 and NIST SD 302a }
\centering
\begin{adjustbox}{width=\textwidth}
\begin{tabular}{cccccccccccccc}
\toprule
\multirow{2}{*}{\textbf{Method}}& \multicolumn{6}{c}{\textbf{\makecell{MI of ROIs Extracted from the Original Fingerprint Pair ($I'_{ro1}$ and $I_{ro2}$)}}}&  & \multicolumn{6}{c}{\textbf{MI of the Enhanced Fingerprint Pair ($I'_{e1}$ and $I_{e2}$)}}\\
\cmidrule(r){2-7}
\cmidrule(r){9-14}

& \textbf{\makecell{FVC \\ 2004 DB1}}& \textbf{\makecell{FVC \\2004 DB2}}& \textbf{\makecell{FVC \\2004 DB3}}&\textbf{MOLF}& \textbf{\makecell{NIST \\ SD4}}& \textbf{\makecell{NIST \\ SD302a}}&  & \textbf{\makecell{FVC \\ 2004 DB1}}& \textbf{\makecell{FVC \\ 2004 DB2}}& \textbf{\makecell{ FVC \\ 2004 DB3}}& \textbf{MOLF}& \textbf{\makecell{NIST \\ SD4}}& \textbf{\makecell{NIST \\SD302a}}\\
 \toprule
 Unaligned&0.42&0.19& 0.40&0.26& 0.15& 0.36&  &0.47& 0.49& 0.69& 0.56& 0.42&0.37\\
 \makecell{Dense Registration in IFViT}& 0.54& \textbf{0.38}& \textbf{0.46}& \textbf{0.33}& \textbf{0.19}& 0.35& & \textbf{0.74}& \textbf{0.90}& \textbf{0.90}& \textbf{0.91}& \textbf{0.83}&\textbf{0.74}\\
 SIFT& 0.24& 0.33& 0.34& 0.14& 0.16& 0.11&  &0.27& 0.31& 0.31& 0.28& 0.22&0.19\\
STN\_DeepPrint& 0.70&0.19& 0.40&0.30& 0.15& 0.34&  &0.65& 0.87& 0.85& 0.85& 0.78&0.72\\
STN\_AFRNet& \textbf{0.71}&0.18& 0.40&0.32& 0.14& \textbf{0.38}&  &0.64& 0.87& 0.84& 0.83& 0.78&0.68\\
\bottomrule
\end{tabular}
\label{table:alignment}
\end{adjustbox}
\end{table*}

\begin{table*}[ht]
\caption{EER of aligned fingerprints in FVC2004 DB1, DB2, DB3, MOLF, NIST SD4 and NIST SD 302a produced from different methods}
\centering
\begin{adjustbox}{width=\textwidth}
\begin{tabular}{cccccccccccccc}
\toprule
\multirow{2}{*}{\textbf{Method}}& \multicolumn{6}{c}{\textbf{\makecell{EER of ROIs Extracted from \\ the Original Fingerprint Pair ($I'_{ro1}$ and $I_{ro2}$)}}}&  & \multicolumn{6}{c}{\textbf{\makecell{EER of the Overlapped Region Extracted from \\ the Enhanced Fingerprint Pair ($I'_{oe1}$ and $I_{oe2}$)}}}\\
\cmidrule(r){2-7}
\cmidrule(r){9-14}

& \textbf{\makecell{FVC \\ 2004 DB1}}& \textbf{\makecell{FVC \\2004 DB2}}& \textbf{\makecell{FVC \\2004 DB3}}&\textbf{MOLF}& \textbf{\makecell{NIST \\ SD4}}& \textbf{\makecell{NIST \\ SD302a}}&  & \textbf{\makecell{FVC \\ 2004 DB1}}& \textbf{\makecell{FVC \\ 2004 DB2}}& \textbf{\makecell{ FVC \\ 2004 DB3}}& \textbf{MOLF}& \textbf{\makecell{NIST \\ SD4}}& \textbf{\makecell{NIST \\SD302a}}\\
 \toprule
 Unaligned&7.75\%&7.38\%& 6.88\%&6.42\%& 7.35\%& 5.49\%&  &5.52\%& 5.32\%& 5.15\%& 7.47\%& 6.82\%&7.00\%\\
\makecell{Dense Registration in IFViT}& \textbf{2.87\%}& \textbf{2.40\%}& \textbf{2.57\%}& \textbf{2.93\%}& \textbf{3.58\%}& \textbf{3.12\%}& & \textbf{2.71\%}& \textbf{2.60\%}& \textbf{2.67\%}& \textbf{3.37\%}& \textbf{4.31\%}&\textbf{2.85\%}\\
 SIFT& 27.42\%& 26.53\%& 28.77\%& 19.33\%& 42.03\%& 23.42\%&  &23.31\%& 16.55\%& 39.66\%& 31.75\%& 44.22\%&25.62\%\\
STN\_DeepPrint& 4.77\%&4.72\%& 4.75\%&5.79\%& 6.63\%& 5.27\%&  &4.88\%& 4.83\%& 4.82\%& 5.77\%& 6.44\%&5.43\%\\
STN\_AFRNet& 4.31\%&4.22\%& 4.27\%&5.28\%& 6.82\%& 5.06\%&  &4.34\%& 4.22\%& 4.23\%& 5.92\%& 6.45\%&5.08\%\\
\bottomrule
\end{tabular}
\label{table:alignmentEER}
\end{adjustbox}
\end{table*}

\subsection {Implementation Details}
In this study, the interpretable dense registration module and the interpretable fixed-length representation extraction and matching module in IFViT are implemented based on Pytorch and trained independently on four NVIDIA GeForce RTX 3090 GPUs. 

In terms of the interpretable dense registration module, a total of 200K fingerprint images (100K fingerprint pair) are used for training. The input images are resized to 128 × 128 to accelerate the convergence process. The module is trained over 100 epochs with a learning rate of  $1 \times 10^{-3} $. For the interpretable fixed-length representation extraction and matching module, we initially pretrain it employing 6.7M synthetic images from PrintsGAN with a learning rate of $1 \times 10^{-3} $. Then, 575K real fingerprint images are used for fine-tuning this module over 70 epochs with a learning rate of $1 \times 10^{-4} $.  Both of these two modules are optimized by Adam with a weight decay of $2 \times 10^{-4} $ and a batch size of 128.

\subsection{Performance of Fingerprint Dense Registration}
Due to the lack of ground truth labels for pixel-wise correspondences of feature points in real fingerprint pairs, it is challenging to evaluate the performance of the fingerprint dense registration algorithm directly through the positions of correspondences. Instead, we execute alignment using correspondences of feature points produced from the registration algorithm and evaluate the alignment performance based on the similarity of aligned enhanced fingerprint pairs ($I'_{e1}$ and $I_{e2}$) and corresponding ROIs ($I'_{ro1}$ and $I_{ro2}$) in original fingerprint pairs. Notably, we do not select overlapped regions from enhanced fingerprint pairs ($I'_{oe1}$ and $I_{oe2}$) for similarity evaluation due to the potential risk of alignment failure, which could result in an extremely limited overlap region. It might lead to similarity computations being based on invalid background areas, finally leading to unreasonable similarity outcomes. 

We select the Mutual Information (MI) metric that has been widely used for image registration in the last decades \cite{maes2003medical, maes1997multimodality} to measure the similarity in aligned fingerprint pairs. Higher MI observed in the fingerprint pair from the same finger indicates superior performance by the registration algorithm. In this study, we compare the alignment performance of the proposed module with two methods: a prevalent open-source algorithm, Scale-Invariant Feature Transform (SIFT), employed for recent registration in fingerprint matching \cite{chowdhury2020can}, and the Spatial Transformer Network (STN), an alignment module widely adopted in many fingerprint recognition tasks \cite{engelsma2019learning, grosz2023afrnet, takahashi2020fingerprint}.  The STNs trained in DeepPrint and AFRNet \cite{engelsma2019learning, grosz2023afrnet} are introduced due to their state-of-the-art performance on fingerprint benchmark datasets.

\begin{figure}
  \begin{center}
  \includegraphics[width=3.3 in]{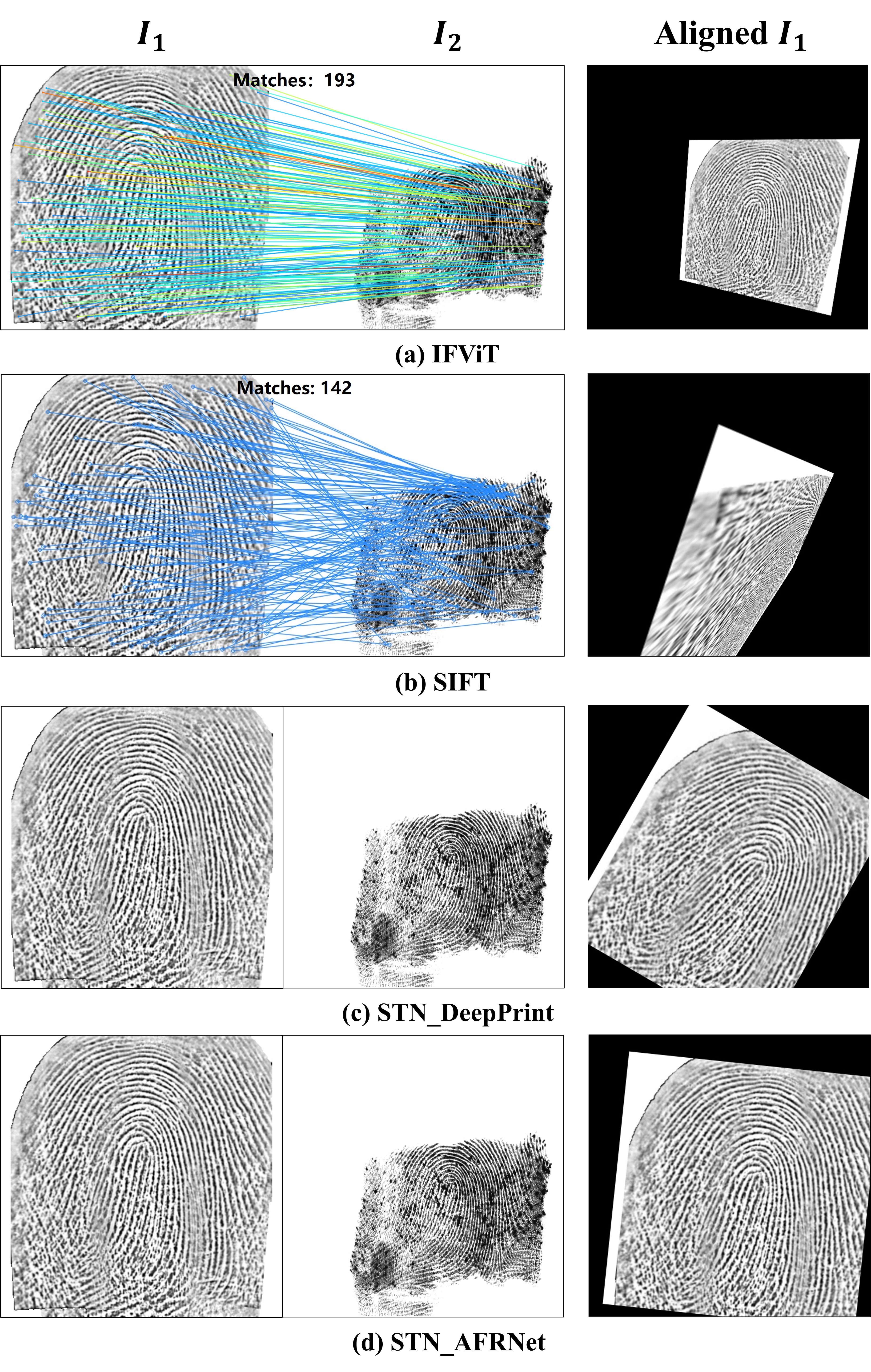}
  
  \caption{{A case of the aligned fingerprint produced by different approaches.}}\label{fig:correspondences}
  \vspace{-0.7cm} 
  \end{center}
\end{figure}

\begin{figure*}
\centering     
\includegraphics[width=7in]{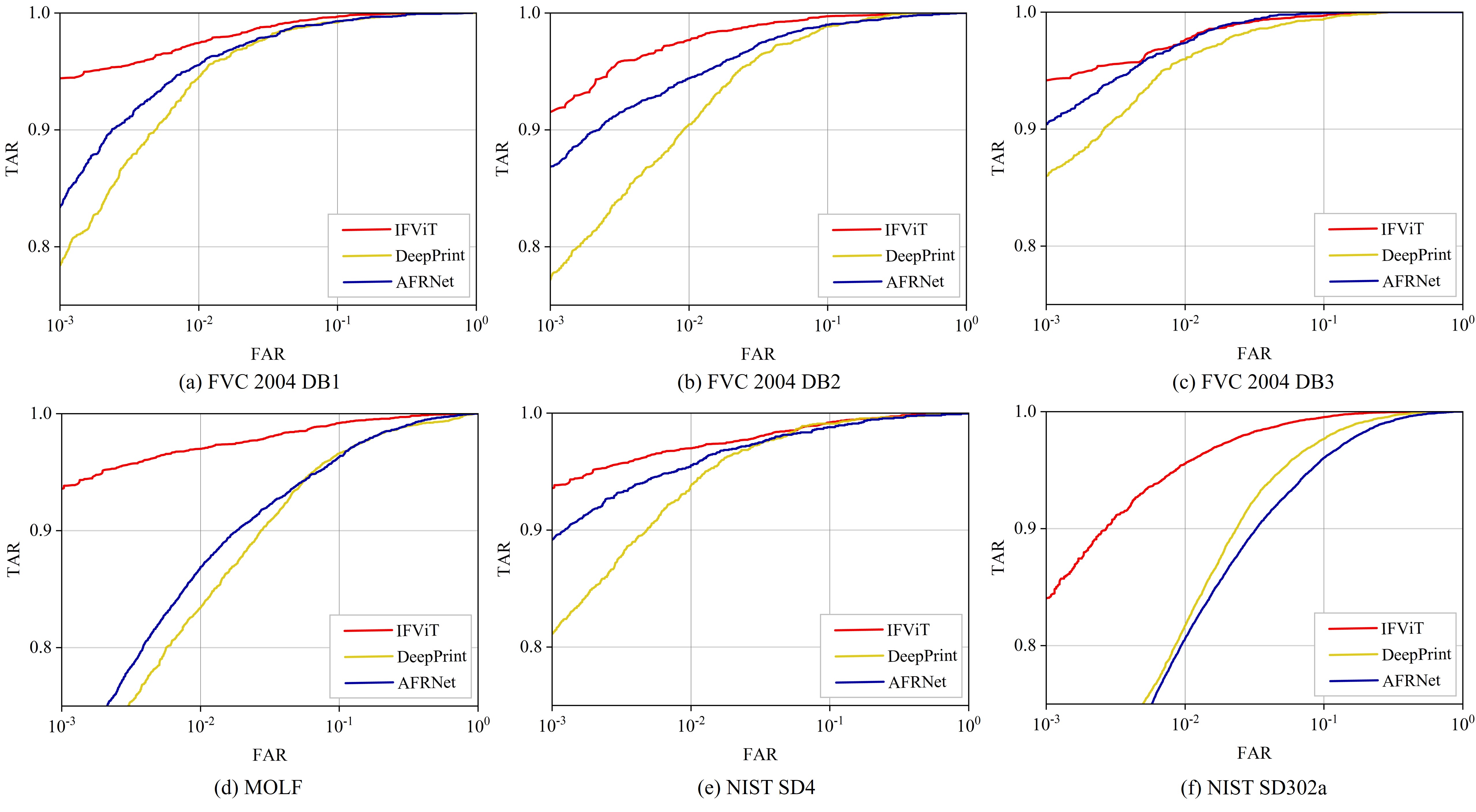}
\centering
\caption{{The ROC curve for (a) FVC2004 DB1 (b) FVC2004 DB2 (c) FVC2004 DB3 (d) MOLF (e) NIST SD4 (f) NIST SD302a.}}\label{fig:ROC}
\end{figure*}

We report MI values based on different methods in Table \ref{table:alignment}. As we can observe 1) The performance of the interpretable dense registration module in the proposed IFViT is more accurate than other methods with better alignment performance. This is evident in its highest MI value on ROIs of original fingerprint pairs and aligned enhanced fingerprint pairs from the same finger across most datasets. 2) The MI values on aligned enhanced fingerprint pairs are generally higher than those on ROIs of original fingerprint pairs. This is primarily attributed to the MI value being influenced by many factors e.g. the quality disparities in original fingerprint pairs, and the variations in fingerprint appearance caused by various sensors.  However, following enhancement by FingerNet, all enhanced fingerprint images adhere to uniform quality standards, reducing their inconsistencies. In this context, the MI value of the latter provides a more precise indication of fingerprint alignment performance. 3) Relative to MI values of unaligned fingerprint pairs, MI values of aligned fingerprint pairs using the SIFT algorithm have shown a marked decline. This is due to the propensity of SIFT to generate incorrect correspondences of feature points, which in turn result in erroneous affine transformations during the alignment procedure. It compromises the structural integrity of original fingerprint images, as illustrated in Fig. \ref{fig:correspondences} (b). 4) The STN in AFRNet and DeepPrint show a comparable level of alignment performance. Notably, STN in AFRNet has superior MI values on ROIs of original fingerprint pairs in the FVC2004 DB1 and NIST SD302a. However, it has slightly worse MI values for enhanced fingerprint pairs compared to the proposed dense registration module. Besides, due to the limitation of STNs in these models to only perform rotation and translation, they cannot achieve pixel-wise alignment for cross-sensor fingerprint pairs, as shown in Fig. \ref{fig:correspondences} (c) (d).

We further explore the influence of fingerprint alignment from various methods on the matching performance based on the trained proposed model, and report their Equal Error Rate (EER) in Table \ref{table:alignmentEER}. The results indicate that the quality of fingerprint alignment significantly impacts the model's matching performance. When considering pixel-wise correspondences of feature points established by the proposed dense registration module, the aligned fingerprints demonstrate the lowest EER in the ROIs from original fingerprint pairs and the overlapped region extracted from aligned enhanced fingerprint pairs across diverse datasets. In comparison to the other methods, our approach is more robust, even for cross-sensor fingerprint pairs, resulting in a considerable number of precise correspondences of feature points between them as shown in Fig. \ref{fig:correspondences} (a).

\subsection{Performance of Fingerprint Matching}

We compare the matching performance of the proposed IFViT with two recent well-known works in fingerprint recognition called DeepPrint and AFRNet \cite{engelsma2019learning, grosz2023afrnet}.  Aside from the minutiae points in the DeepPrint network, which are generated by FingerNet as an alternative for creating minutiae maps, the employed model structures are as consistent as possible with those reported in their studies.

From the EER reported in Table \ref{table: EER} and the full Receiver Operating Characteristic (ROC) shown in Fig. \ref{fig:ROC}, it can be observed that the proposed IFViT outperforms the other two approaches across six benchmark databases, achieving the lowest EER. The reason for this discrepancy is that we observe a decrease in the performance of DeepPrint and AFRNet when there is an imbalance in the quantity of  fingerprints collected from different manners (e.g., rolled, plain, and slap) on the training set. In the context of a limited number of fingers, the developed IFViT can consider both the ROIs of the original fingerprint pairs and the overlapped regions from enhanced fingerprint pairs to reduce the disparity among fingerprint images, thus improving the model's generalization capacity.  Particularly with cross-sensor fingerprints from the MOLF and NIST SD302a, predicting them poses a challenge for both DeepPrint and AFRNet. But for IFViT, it can notably reduce the EER by integrating the dense registration module for pixel-wise precise alignment and combining local and global representations within the matching module. 


\begin{table}[ht]
\caption{EER in FVC2004 DB1, DB2, DB3, MOLF, NIST SD4 and NIST SD302a based on diverse matching methods}
\centering
\begin{adjustbox}{width=3.5in}
\begin{tabular}{ccccccc}
\toprule
\multirow{2}{*}{\textbf{Method}}& \multicolumn{6}{c}{\textbf{EER}}\\
\cmidrule(r){2-7}

& \textbf{\makecell{FVC \\ 2004 DB1}}& \textbf{\makecell{FVC \\2004 DB2}}& \textbf{\makecell{FVC \\2004 DB3}}&\textbf{MOLF}& \textbf{\makecell{NIST \\ SD4}}& \textbf{\makecell{NIST \\ SD302a}}\\
 \toprule
 IFViT&\textbf{1.87\%}&\textbf{1.83\%}& \textbf{1.49\%}&\textbf{2.33\%}& \textbf{2.71\%}& \textbf{2.31\%}\\
 DeepPrint& 2.70\%& 3.39\%& 2.05\%& 5.18\%& 2.85\%& 5.03\%\\
AFRNet& 2.43\%&2.95\%& 1.56\%&5.33\%& 2.80\%& 6.29\%\\
\bottomrule

\end{tabular}
\end{adjustbox}
\label{table: EER}
 \vspace{-0.5cm} 
\end{table}

\begin{figure*}
  \begin{center}
  \includegraphics[width=7in]{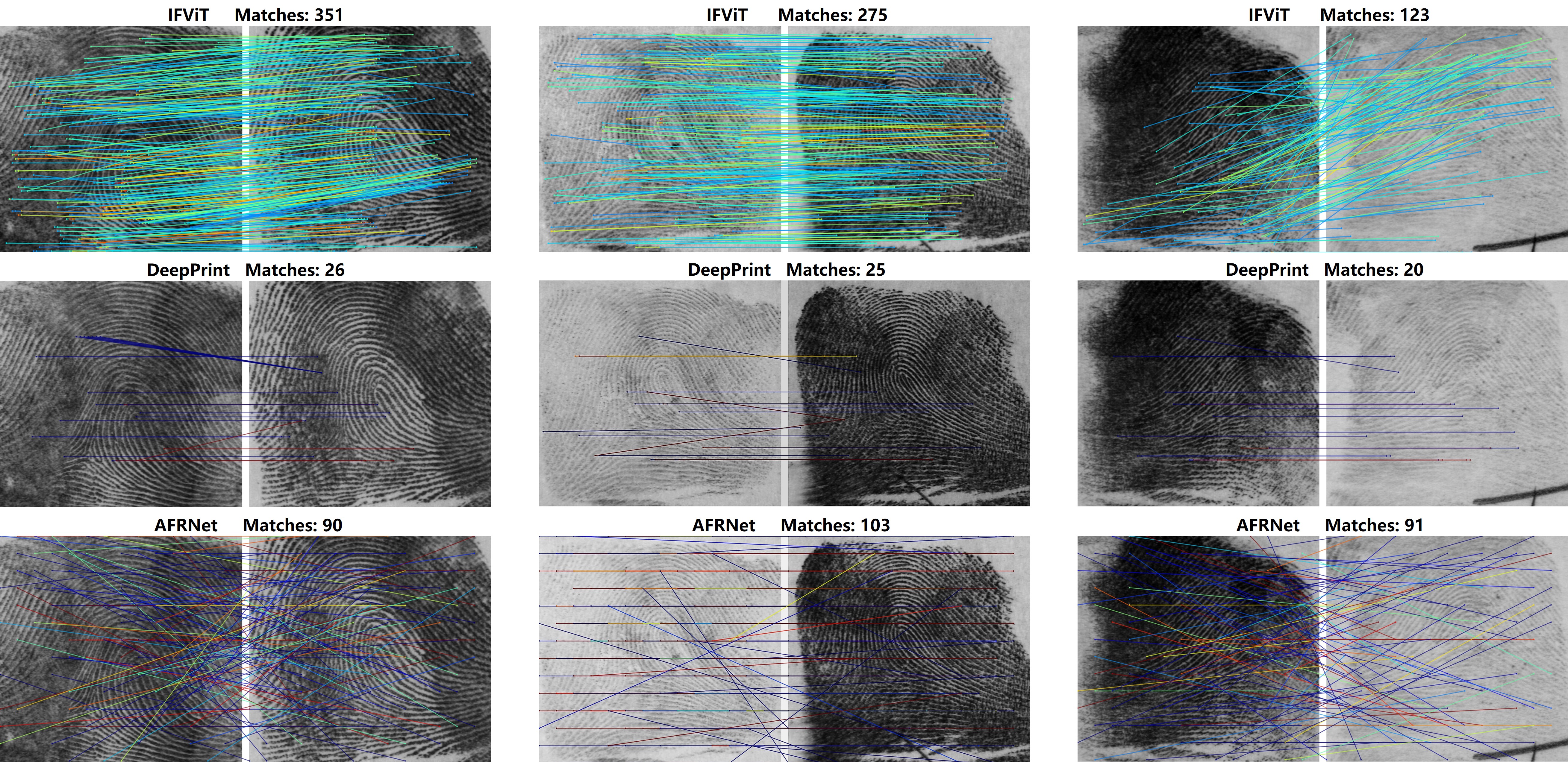}
  \caption{{Cases of interpretable pixel-wise correspondences of feature points produced from IFViT, DeepPrint and AFRNet based on the fingerprint pair from the same identity.}}\label{Interpretability1}
  \end{center}
  \vspace{-0.6cm} 
\end{figure*}

\subsection{Interpretability}
In addition to demonstrating the interpretability of fingerprint alignment shown in Fig. \ref{fig:correspondences}. We further investigate the interpretability of the proposed matching module. Based on the implementation reported by AFRNet \cite{grosz2023afrnet}, we evaluate the interpretability of AFRNet by computing the correspondences of feature points in the fingerprint pairs. In addition, similar operations are also performed on the CNN output of DeepPrint \cite{engelsma2019learning} to compare with our method on original fingerprint images. As shown in Fig. \ref{Interpretability1}, when matching fingerprint pairs from the same identity, the matching module in the proposed IFViT  can offer more reasonable and superior interpretability. DeepPrint can only focus on local regions and establish limited number of correspondences. AFRNet, due to its adoption of self-attention in ViT that can capture long-range dependencies and global context, similar to the proposed IFViT, can provide more correspondences in the global context, but it is challenging to provide reasonable ones based on the matching result. 

\begin{figure}
  \begin{center}
  \includegraphics[width=3.5in]{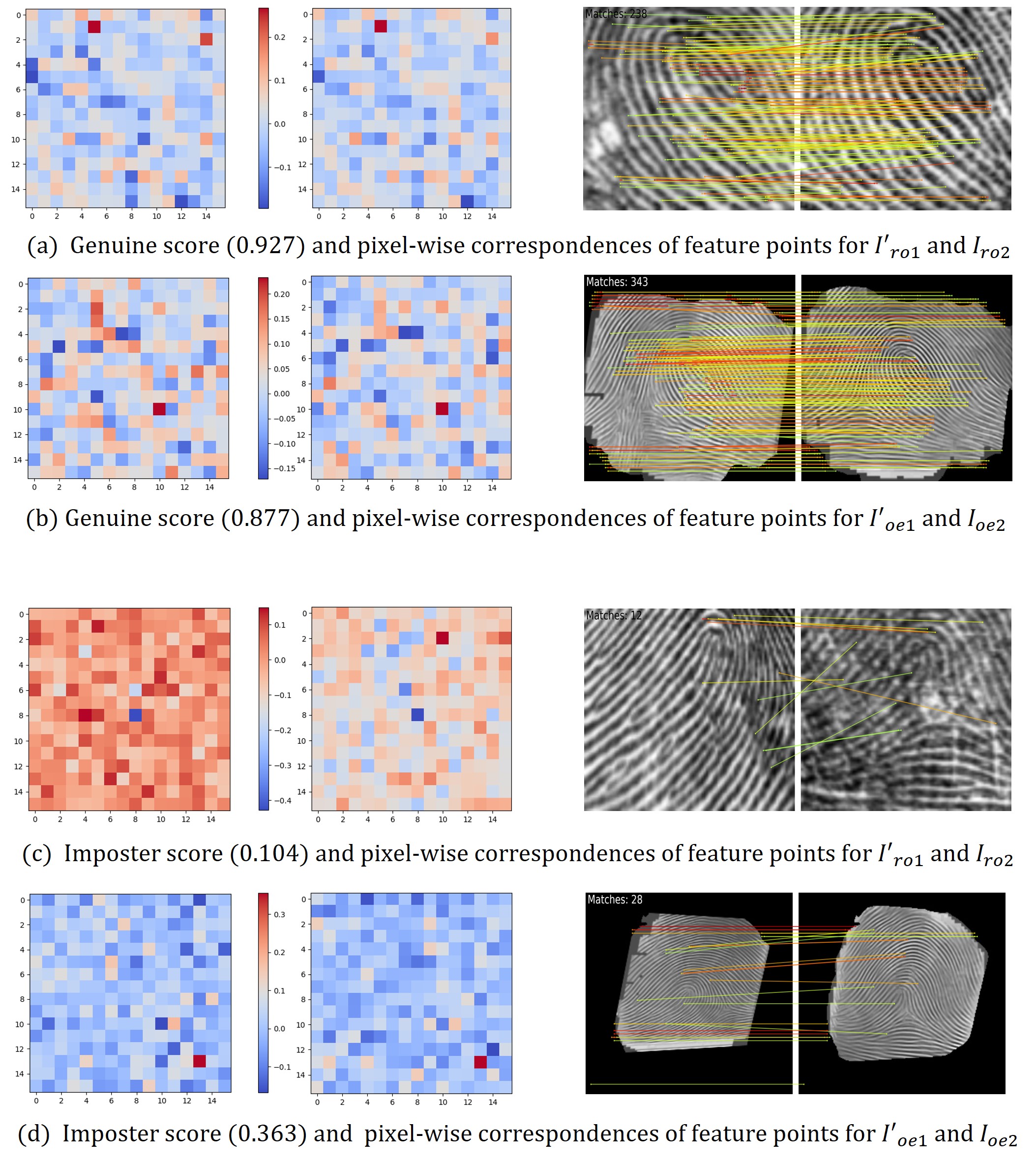}
  \vspace{-0.3cm} 
  \caption{{Illustration of IFViT interpretability. The learned 256-dimensional fixed-length fingerprint deep representation is shown as the heapmap, along with corresponding interpretable pixel-wise correspondences of feature points in genuine pair (a) (b) and imposter pair (c) (d). 
  \vspace{-0.9cm} 
}}\label{fig: interpretability}
  \end{center}
\end{figure}

Fig. \ref{fig: interpretability} further illustrates the interpretability of IFViT during the matching procedure by using the fusion of local and global representations. It can calculate the matching score for the given fingerprint pair while simultaneously establishing interpretable dense pixel-wise correspondences of feature points across both the ROI extracted from the original fingerprint ($I_{ro1}'$ and $I_{ro2}$) and the overlapped region from the enhanced fingerprint ($I_{oe1}'$ and $I_{oe2}$) in the matching result. This aids in understanding the reasons behind the decision made by the deep learning model during fingerprint matching. In the example provided in Fig. \ref{fig: interpretability}, the imposter pair establishes fewer correspondences of feature points during prediction compared to the genuine pair, with the vast majority of them being incorrect.

\begin{figure}
  \begin{center}
  \includegraphics[width=3.4in]{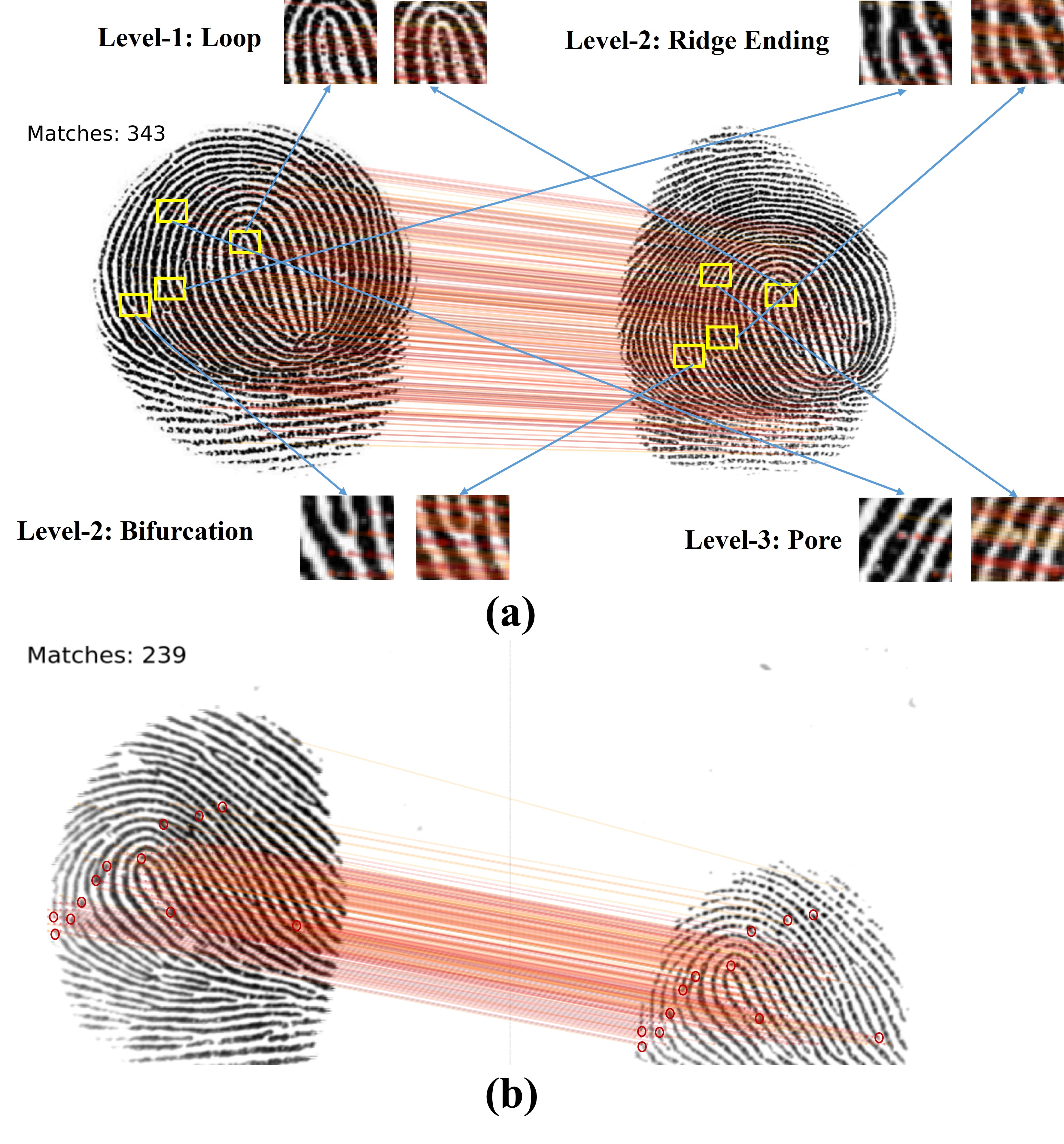}

  \caption{{Examples of feature points highlighted in established dense correspondences for the given fingerprint pair from the same identity with a confidence score exceeding 0.7: (a) Highlighted feature points of the fingerprint at different levels (b) Highlighted minutiae.}}
  \vspace{-0.3cm}
  \label{fig: minutiaepoints}
 
  \end{center}
   \vspace{-0.3cm} 
\end{figure}

To further observe which feature points of fingerprints are possibly used in fixed-length representation-based matching, we provide examples of learned correspondences with a high confidence level. Fig. \ref{fig: minutiaepoints} (a) showcases that feature points of fingerprints at different levels can be identified around learned dense correspondences by IFViT. These features range from the global-level feature (level-1) such as the loop, to local-level features (level-2) represented by minutiae composed of ridge endings and bifurcations, down to the very fine-level feature (level-3) typified by the pore. This implies that, although we do not explicitly integrate domain knowledge into the model, to some extent, the model can infer it on its own. Moreover, we also discover that as the resolution of fingerprint images decreases, such as in cases where pores cannot be clearly distinguished, or when the matching region is limited, there is a noticeable shift towards emphasizing the minutiae in learned dense correspondences as shown in Fig. \ref{fig: minutiaepoints} (b). This is aligned with our understanding that minutiae are one of the most important features adopted in fingerprint matching algorithms. 

Instead of using techniques such as Grad-CAM or Saliency Map \cite{chowdhury2020can,takahashi2020fingerprint}, which rely on gradient computation to identify general regions contributing to predictions in the input fingerprint pair, IFViT can provide superior interpretability at the pixel level. It allows for the detailed examination of specific fingerprint feature points that are used and transformed into a fixed-length representation for matching, offering insights into the decision-making process of black-box deep learning models for both successful and failed fingerprint matching cases. 

\vspace{-0.3cm}
\subsection{Ablation Study}
\begin{table*}[ht]

\caption{The ablation study of the proposed IFViT}
\centering
\begin{adjustbox}{width=7in}
\begin{tabular}{cccccccccccc}
\toprule
\multirow{2}{*}{\textbf{Method}}&      \multirow{2}{*}{\textbf{Alignment}} & \multirow{2}{*}{\textbf{\makecell{Local \\ Representation}}}& \multirow{2}{*}{\textbf{\makecell{Global \\ Representation}}}& \multirow{2}{*}{\textbf{ \makecell{Fusion of \\ Local and  Global Representation}}}&&\multicolumn{6}{c}{\textbf{EER}}\\
\cmidrule(r){7-12}

&      &&&&&\textbf{\makecell{FVC \\ 2004 DB1}}& \textbf{\makecell{FVC \\2004 DB2}}& \textbf{\makecell{FVC \\2004 DB3}}&\textbf{MOLF}& \textbf{\makecell{NIST \\ SD4}}& \textbf{\makecell{NIST \\ SD302a}}\\
 \toprule
 Proposed Method (A)&     &\checkmark&&&&47.63\%&46.40\%& 47.07\%&36.80\%& 35.36\%& 51.63\%\\
Proposed Method (B)&      &&\checkmark&&&31.16\%& 30.14\%& 29.66\%& 42.52\%& 12.92\%& 39.69\%\\
 Proposed Method (C)&      \checkmark&\checkmark&&&&2.87\%& 2.40\%& 2.57\%& 2.93\%& 3.58\%& 3.12\%\\
 Proposed Method (D)& \checkmark& & \checkmark& & & 2.71\%& 2.60\%& 2.67\%& 3.37\%& 4.31\%&2.85\%\\
Proposed Method (E)&      \checkmark&\checkmark&\checkmark&\checkmark&& \textbf{1.87\%}& \textbf{1.83\%}& \textbf{1.49\%}&\textbf{2.33\%}& \textbf{2.71\%}& \textbf{2.31\%}\\
\bottomrule
\end{tabular}
\end{adjustbox}
\label{table:ablation study}
\end{table*}
To justify the incorporation of each component (i.e. fingerprint alignment, local representation and global representation), we conduct an ablation study by sequentially introducing each module into the matching pipeline. The outcomes, as shown in Table \ref{table:ablation study}, demonstrate improved matching performance with the addition of each component. It highlights the importance of the proposed alignment module in improving fingerprint matching performance, especially in cases involving fingerprint pairs with limited overlapping areas. Furthermore, the combination of local and global representations can compensate for the deficiencies in semantic information present in each, thereby achieving an improved recognition rate.
\vspace{-0.2cm} 
\subsection {Computational Efficiency}
The inference time with the proposed IFViT primarily involves three stages: fingerprint enhancement by FingerNet, registration and alignment, and matching. Running on an Intel(R) Xeon(R) Gold 6133 CPU @ 2.50GHz with the NVIDIA GeForce RTX 3090 GPU, it takes approximately 463ms for inference based on a given fingerprint pair (116ms for enhancement, 187ms for registration and alignment, and 160ms for matching).  To further improve the computational efficiency, the deep convolutional neural network (i.e. ResNet-18) for local feature extraction in IFViT could be replaced by other shallower networks. Also, the FingerNet could be substituted with alternative fingerprint enhancement techniques with lower computational complexity. We do not consider optimization of them, as this falls outside the scope of this study.

\vspace{-0.1cm}
\section {Conclusion}
In this paper, we propose a novel multi-stage interpretable fingerprint matching network via vision transformer called IFViT. This is the first attempt to utilize the ViT to achieve fingerprint dense registration. The proposed interpretable fingerprint registration module can generate dense pixel-wise correspondences of feature points and perform precise alignment for the input fingerprint pair. The interpretable fixed-length representation extraction and matching module, achieves lower EER and higher recognition accuracy across multiple datasets when the model is trained on a limited number of fingers, simultaneously demonstrating superior interpretability. In our future work, we plan to further optimize the existing framework and explore its applicability in more challenging scenarios e.g. contactless fingerprints and latent fingerprints to enhance their interpretability during the matching process. 



%



\vspace{-0.15cm}
\section*{Acknowledgment}
The authors are grateful to Zhiyu Pan from Tsinghua University for his valuable suggestions and insights that helped us to improve the quality of this work.


\ifCLASSOPTIONcaptionsoff
  \newpage
\fi




{\footnotesize
\bibliographystyle{IEEEtran}
\bibliography{IEEEabrv,Bibliography}}

\vfill


\end{document}